\documentclass[review,12pt,authoryear]{elsarticle}

\usepackage{amssymb}
\usepackage{tabularx}
\usepackage{graphicx}
\usepackage{longtable}
\usepackage{xltabular}
\usepackage{placeins}
\usepackage{url}

\journal{Engineering Applications of Artificial Intelligence}

\begin{document}

\begin{frontmatter}



\title{B-SMART: A Reference Architecture for Artificially Intelligent Autonomic Smart Buildings}


\author[1]{Mikhail Genkin}
\author[1]{J.J. McArthur}

\affiliation[1]{organization={Department of Architectural Science},
            addressline={Toronto Metropolitan University}, 
            city={Toronto}, 
            state={Ontario},
            country={Canada}}

\begin{abstract}
The pervasive application of artificial intelligence and machine learning algorithms is transforming many industries and aspects of the human experience. One very important industry trend is the move to convert existing human dwellings to smart buildings, and to create new smart buildings. Smart buildings aim to mitigate climate change by reducing energy consumption and associated carbon emissions. To accomplish this, they leverage artificial intelligence, big data, and machine learning algorithms to learn and optimize system performance. These fields of research are currently very rapidly evolving and advancing, but there has been very little guidance to help engineers and architects working on smart buildings apply artificial intelligence algorithms and technologies in a systematic and effective manner. In this paper we present B-SMART: the first reference architecture for autonomic smart buildings. B-SMART facilitates the application of artificial intelligence techniques and technologies to smart buildings by decoupling conceptually distinct layers of functionality and organizing them into an autonomic control loop. We also present a case study illustrating how B-SMART can be applied to accelerate the introduction of artificial intelligence into an existing smart building.
\end{abstract}

\begin{graphicalabstract}
\includegraphics{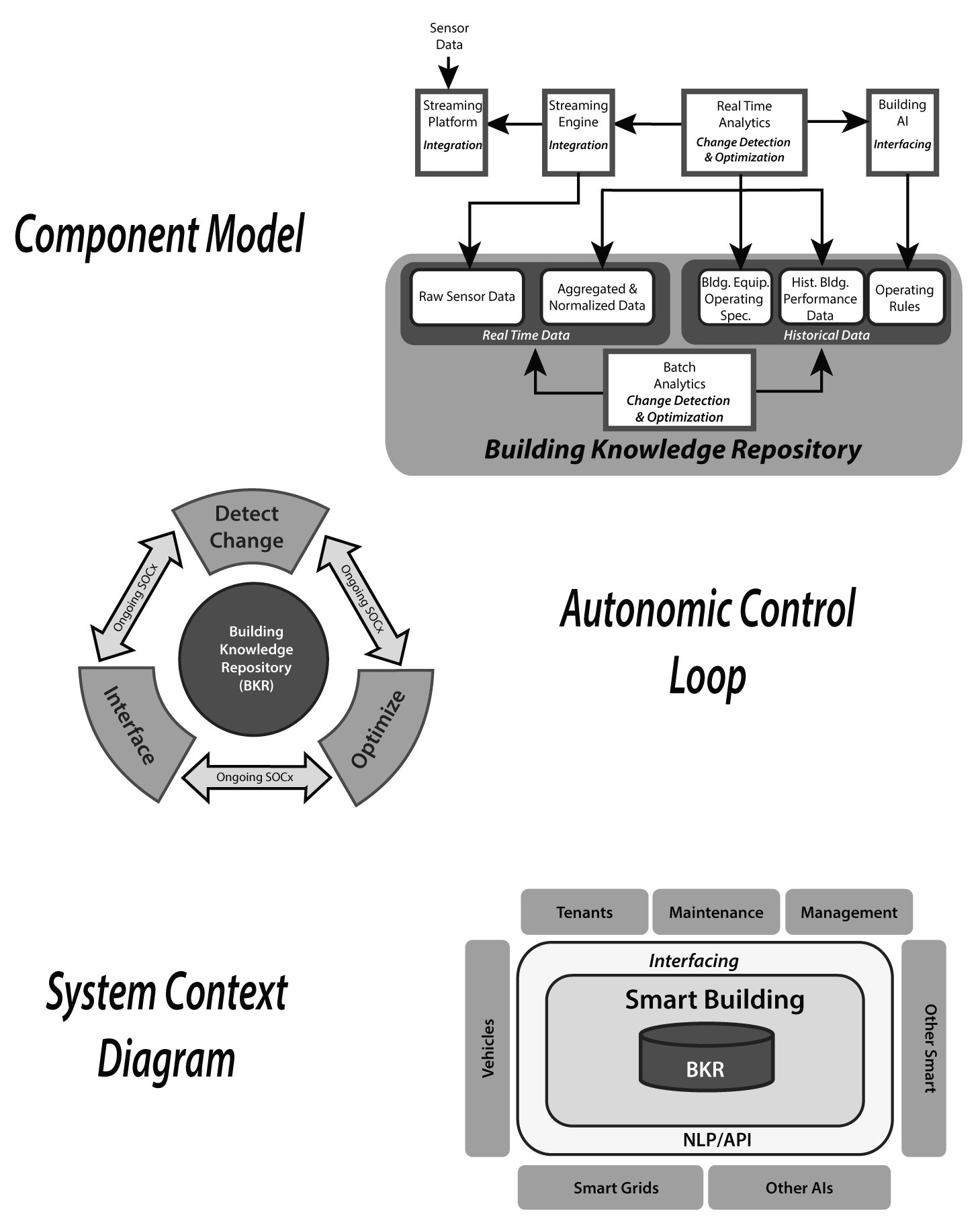}
\end{graphicalabstract}

\begin{highlights}
\item B-SMART is a reference architecture that accelerates the application of artificial intelligence in smart buildings
\item B-SMART supports autonomic smart building operations
\item B-SMART supports both startup and ongoing commissioning
\end{highlights}

\begin{keyword}
artificial intelligence \sep smart buildings \sep reference architecture \sep autonomic computing



\end{keyword}

\end{frontmatter}


\section{Introduction} \label{intro}

Buildings are responsible for approximately 32\% of global energy use and 19\% of CO$_2$ emissions and their longevity as well as the ability to reduce these values have made it a significant priority for emissions reduction \citep{IPCC2018}. In this context, the ability to manage buildings most efficiently is critical. Significant research and best practice development have formed a body of commissioning practices implemented throughout the building life-cycle that can reduce energy consumption by 15-30\% \citep{rothetal2008}. Recent developments in machine learning and cloud computing provide new opportunities to implement Smart and Ongoing Commissioning (SOCx); preliminary field studies have demonstrated this approach to provide savings as high as 70\% \citep{stocketal2021}. However, there is limited uptake due to the lack of a supporting computational architecture to integrate machine learning and artificial intelligence with building systems. This paper seeks to address this research gap, building upon the state-of-the-art to present the first reference architecture for autonomic smart buildings to enable SOCx throughout the building life-cycle. The Building Systems Management Autonomic Reference Template (B-SMART) can be used to dramatically accelerate the design of information systems, reduce the building energy footprint, and facilitate the application of Artificial Intelligence (AI) in smart buildings.

\subsection{Autonomic Systems} \label{auto-systems}

In the key technical white-paper outlining their vision of autonomic computing \citep{ibm2005architectural}, IBM introduced the following definition of autonomic computing: "\textit{A computing environment with the ability to manage itself and dynamically adapt to change in accordance with business policies and objectives. Self-managing environments can perform such activities based on situations they observe or sense in the IT environment rather than requiring IT professionals to initiate the task. These environments are self-configuring, self-healing, self-optimizing, and self-protecting.}".

This initial definition was fairly narrow, and since then significant research was done to expand the set of autonomic properties that need to be supported (discussed further in Sub-section ~\ref{backgr-early}). Also, autonomic computing environments and systems do not exist in a void. They must interact with other actors in their environment, and so in this work we define an autonomic cybernetic system or computing environment as follows: "\textit{An autonomic cybernetic system is a computing environment that is capable of recognizing and responding to change in it's internal operating characteristics or external operating environment with minimal human intervention, and capable interfacing with other humans or autonomic systems in accordance with business objectives, rules or policies.}"

\subsection{Smart Buildings and SOCx} \label{intro-smt-bld-socx}

The term \textit{Smart Buildings} was initially defined as those buildings whose "design and construction require the integration of complex new technologies into the fabric of the building” \citep{dreweretal1994}. In the intervening decades, several new developments have expanded the definition of Smart Buildings, but at the core remains this need for integration of a diversity of new technologies including those yet to be developed. This anticipation of future learning has led to significant applications of artificial intelligence and machine learning within this domain. Driven by the urgency of climate change, energy management has been a particular focus for this research with sub-focuses on monitoring, fault detection and diagnosis, and scheduling problems \citep{aguilar2021}. Advances in Building Information Modeling (BIM) have been developed to integrate the information required to support such Smart Building applications throughout the building life-cycle \citep{panteli2020}. Complementing this, significant research has developed solutions for the streaming of Building Automation System (BAS) and other sensor network data \citep{misicetal2020}. However, legacy BAS systems pose a significant challenge to this integration. Further complicating this challenge is the ever-evolving diversity of Internet of Things (IoT) devices and applications enabling Smart Building operations \citep{jia2019, sharmaetal2018}.

Within this context of complex, heterogeneous systems integrated into Smart Buildings, the ongoing optimization of building performance is a significant challenge. SOCx is the integration of traditional commissioning processes with online monitoring and data analysis drawn from IoT devices and traditional building systems such as BAS to maintain optimum building performance \citep{gilani2020, noyeetal2016, minolietal2017}. These SOCx systems will benefit building operators and facility managers in three key ways: (1) a reduction in nuisance alarms and their automatic correction; (2) new insights into fault detection, including their resolution when human intervention is not required; (3) and improved energy performance of the facility. Recognizing the diversity of data required to support these practices, several SOCx ontologies have been developed, as have data streaming approaches to collect these data \citep{misicetal2020} and algorithms to support fault detection and energy optimization \citep{MARIANOHERNANDEZ2021101692}. Despite this innovation, however, there remains a paucity of literature regarding autonomic smart building approaches. Few autonomic architectures for smart buildings exist, and those currently representing the state-of-the-art fail to fully address the needs of SOCx processes as they lack key autonomic system properties. In addition, there remains a heavy reliance on manual data collection, transformation, and analysis to populate them, which are slow and expensive processes. This is a significant factor slowing smart building adoption. Ideally a smart building, like advanced Industrial IoT systems \citep{koziolek2018}, should be able to commission itself, re-commission itself if the situation calls for it, and maintain optimal systems performance on an on-going basis.

The future of smart buildings is autonomic. To overcome the barriers to implementation discussed above, we propose the first reference architecture developed specifically to address the needs of smart buildings. Our architecture supports all the key properties of autonomic systems and is tailored specifically to support SOCx implementation in Smart Buildings. Through this contribution, we seek not only to enable their implementation, but to provide a guide for continuing research and development activities in this very important area.

This paper is laid out as follows. First, we present a review of existing autonomic architectures for smart buildings, contextualized within the broader development of autonomic computing. Next, we outline our methodology, which draws from this literature review to develop the requirements for reference architecture development. In Sections ~\ref{comp-arch}, ~\ref{layering}, and ~\ref{ctrl-loop}, we present the B-SMART – the first reference architecture for autonomic smart buildings. In Section ~\ref{basintegration} we discuss how legacy BAS can be incorporated into B-SMART. In Section ~\ref{supporting-socx} we discuss how B-SMART supports the SOCx processes. In Section ~\ref{case-study} we present an example that illustrates how B-SMART can be applied on an existing smart building to help plan the development of additional smart building features.  Finally, we discuss the applications and implications for this reference architecture and conclude with a summary of our findings, the limitations of this work, and recommendations for future research.

\section{Background} \label{background}

Recent advances in AI and Machine Learning (ML) have triggered evolutionary changes across many industries known as Industry 4.0. Examples include self-driving cars, autonomous drones, robotic production lines, AI assisted medical diagnosis, social network algorithms, facial-recognition features in smart phones and cameras, and many other examples. Such modern autonomic systems pervasively apply ML algorithms to identify objects that collectively comprise their environment and classify them into useful categories. Once detected and classified, the autonomic system needs to decide what actions need to be performed on, or in response to, those objects. The actions to be performed could be either learned using ML algorithms or selected using a rules-based system.

Autonomous systems typically need to co-exist and interact with humans. This interaction can be either \textit{human-in-the-loop} and \textit{human-on-the-loop}. Human-in-the-loop systems typically identify and classify the objects in their environment, present a human operator with a proposed set of actions, and wait for the human to confirm which action should be taken. Human-on-the-loop systems provide the human operator with the same level of information but do not wait for human operator to confirm; instead, the action will be performed automatically. The human operator will be able to observe and, if necessary, intervene in the autonomic cycle.

\subsection{Early Autonomic Computing Architectures} \label{backgr-early}

The field of autonomic computing was introduced by IBM in 2001 with the goal of creating computer systems capable of self management. Since then, there has been considerable research focusing on developing autonomic systems. Originally IBM defined autonomic systems to exhibit the following characteristics:
\begin{enumerate}
\item \textbf{\textit{Self-configuration}}. The ability to configure its own components without human
intervention.
\item \textbf{\textit{Self-healing}}. The ability to recognize and correct faults without human intervention.
\item \textbf{\textit{Self-optimization}}. The ability to monitor and optimize their own performance.
\item \textbf{\textit{Self-protection}}. The ability of the system to detect intrusion and defend itself from its
unwanted effects.
\end{enumerate}

\begin{figure}
\begin{center}
\includegraphics[width=0.8\textwidth]{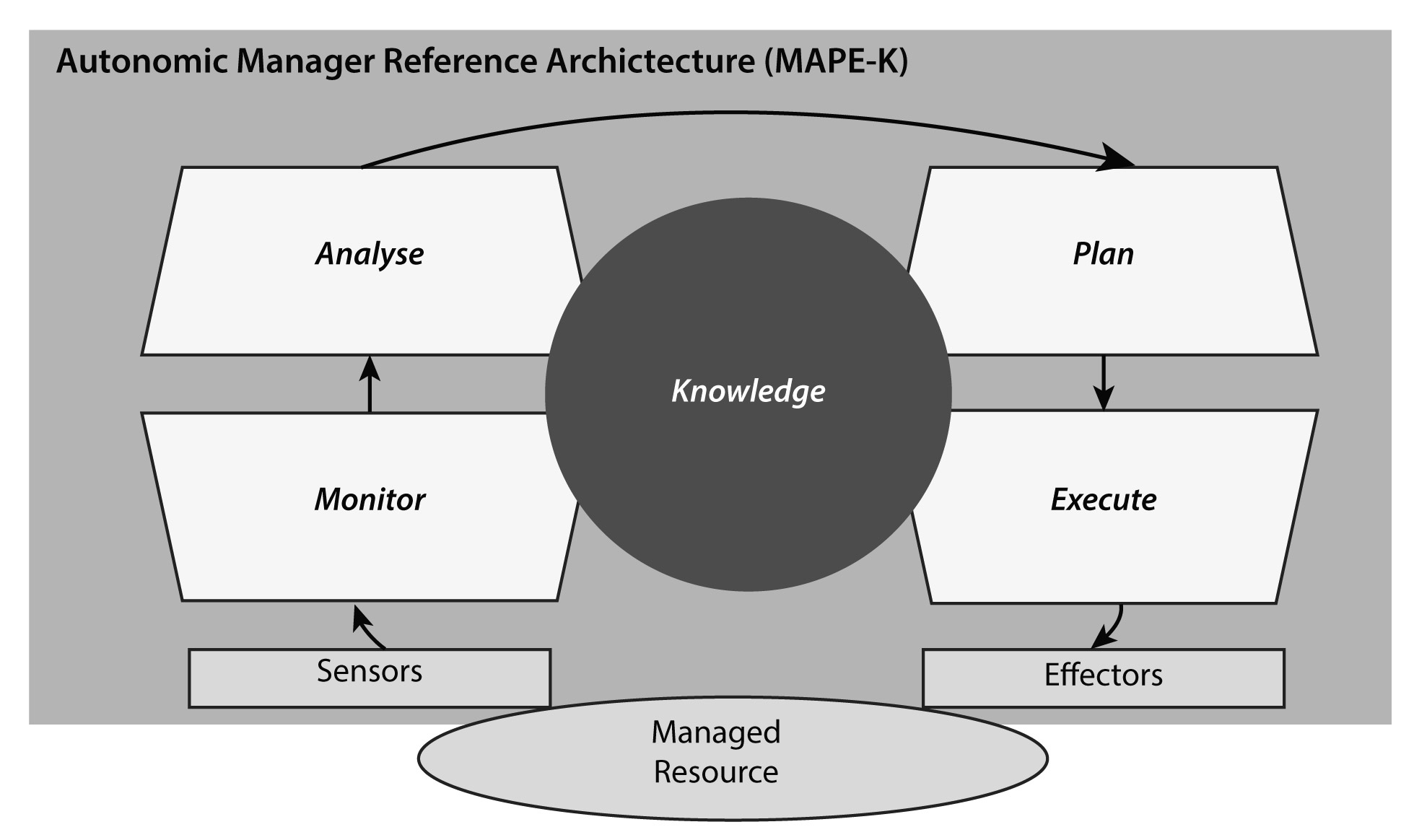}
\end{center}
\caption{\label{mape-k} The IBM MAPE-K reference architecture for autonomic managers \citep{ibm2005architectural}.}
\end{figure}

In 2005 IBM published its reference architecture for autonomic systems \citep{ibm2005architectural}. This reference architecture, referred to as MAPE-K, is shown in Figure ~\ref{mape-k}. The acronym MAPE-K stands for Monitor (M), Analyze (A), Plan (P), Execute (E), and Knowledge (K).

The IBM MAPE-K reference architecture remains the most widely used autonomic reference architecture today. For example, it is referenced in all works cited in Table 1 focusing on big data, networking, and cloud computing; note that the cited works focusing on the smart building space are not really autonomic architectures. It was observed that key to the definition of autonomic systems was the notion of supporting certain autonomic system properties. MAPE-K is a conceptual architecture that does not provide guidance on how to effectively leverage the algorithms, techniques and technologies from the rapidly evolving AI domain when designing smart buildings. Smart building developers would need to acquire deep AI skills and define a more domain-relevant architecture, complete with a technology mapping. This is expensive and time-consuming. To address this issue our autonomic reference architecture for smart buildingsuses the MAPE-K reference architecture as a starting point, and enhances it with domain specific insight.

Other researchers \citep{nami2007, proslad2009} have expanded these desirable characteristics to add:

\begin{enumerate}
\item \textbf{\textit{Self-regulation}}. The ability to maintain a key parameter, such as the quality of service.
\item \textbf{\textit{Self-learning}}. The ability to learn about its environment without external intervention.
\item \textbf{\textit{Self-awareness}}. The ability to be aware of its constituent components and key external dependencies.
\item \textbf{\textit{Self-organization}}. The ability to the system to organize its own structure.
\item \textbf{\textit{Self-creation/Self-assembly/Self-replication}}. The ability to organize itself in response to changing strategic goals.
\item \textbf{\textit{Self-Management/Self-Governance}}. The ability to manage all its constituent components.
\item \textbf{\textit{Self-description/Self-representation}}. Humans should be able to understand an autonomous system.
\end{enumerate}

The MAPE-K architecture does not explicitly explain how these autonomic properties should be implemented. Most autonomic architectures listed in Table ~\ref{review-table} focus on implementing the self-optimization property. The other properties are not nearly as widely discussed in currently available scientific literature.

\subsection{Autonomic Architecture for Smart Buildings} \label{backgr-smt}

There have been very few published studies focusing on how to make smart buildings autonomic. None of them describe an autonomic reference architecture. \cite{chevallier2020} present a reference architecture for a smart building digital twin, which is used to model physical building performance. Their reference architecture focuses on how to collect, store, and expose the static and dynamic information produced by smart buildings for query by human operators. Their reference architecture does not specifically address the autonomic aspect of smart building operation, nor does it discuss the implementation of autonomic properties, encouraging instead human interaction with the digital twin.

\cite{bashir2020} describe a conceptual framework called Integrated Big Data Management and Analytics (IBDMA), a reference architecture, and a metamodel for smart buildings. The meta-model describes the different conceptual entities and processes that comprise the smart building, and how they interact with each other. The reference architecture describes how the big data stack technologies can be used to implement these processes. but do not discuss the autonomic properties, nor how these can be implemented in smart buildings.

\cite{mazzara2019} also propose a reference architecture for smart and software defined buildings. Their reference architecture also does not specifically focus on the autonomic aspects of managing smart buildings. It does include a layered view of smart building technologies. The key layers include: Hardware; Network; Management; and Application and Service. The study also discusses automating some of the smart building features but does not describe or discuss the implementation of autonomic properties by their architecture layers. Their architecture targets the more conventional smart building that relies on human interaction rather than autonomic operation and is supplemented by a rules engine to manage and optimize building operation.

\cite{aguilar2019} propose a self-managing architecture for multi-HVAC systems in buildings based on an “Autonomous cycle of Data Analysis Tasks” (ACODAT) concept. While their work is a significant step towards defining an autonomic architecture for smart buildings, it has several shortcomings. Their architecture does not explain layering, or separation of responsibilities among the different technologies. It also does not support several key autonomic systems properties – such as self-organization and self-creation. Further, this study focuses on the optimization aspect of a smart building already in operation, but do not engage with questions of new implementation where there will be no historical data to work with, and thus there must be a lengthy period when it will not be possible to optimize the building power consumption.

\FloatBarrier
\begin{table}[h]
\caption{\label{review-table} Review of Autonomic Architectures Applicable to Smart Buildings.}
\begin{center}
\begin{tabularx}{\textwidth}{X|X|X}
\hline
Architecture&Domain&Supported Autonomic Properties\\
\hline
ACODAT \citep{aguilar2019, aguilar2021}&Smart Buildings&Self-optimization\\
\cite{chevallier2020}&Smart Buildings&None\\
SSDB \citep{mazzara2019}&Smart Buildings&None\\
IBDMA \citep{bashir2020}&Smart Buildings&None\\
\cite{qin2014}&Smart Grid/Cloud&Self-optimization\\
PSP \citep{elnaffar2009}&Database&Self-optimization \newline Self-configuration\\
KERMIT \citep{genkin2019, genkin2020}&Big Data&Self-optimization \newline Self-configuration\\
ANA \citep{bouabene2009}&Networking&Self-configuration\\
CAN \citep{elsawy2015}&Networking&Self-configuration \newline Self-optimization\\
VANET \citep{tomar2018}&Cloud Computing&Self-optimization\\
\cite{gergin2014}&Cloud Computing&Self-optimization \newline Self-healing\\
\hline
\end{tabularx}
\end{center}
\end{table}
\FloatBarrier

In summary, most prior works focus on implementing the self-configuration and self -optimization autonomic properties. The MAPE-K autonomic reference architecture, and especially the autonomic control loop described in it, is almost exclusively used. All the reviewed architectures had a notion of technology layering. Autonomic architectures can be either centralized or distributed. Many autonomic architectures leverage ML algorithms and AI techniques that are often grouped into a separate cognitive layer which, among other things, is responsible for detecting change and updating the models. These observations have been reflected in the design of the B-SMART reference architecture, as discussed in the following sections.

\section{Methodology} \label{methodology}

A reference architecture is typically created as a digest of existing architectures. It encapsulates the best practices learned by implementing architectures in the field and serves as a template for new architectures. The objective of the reference architecture is to facilitate creation of new architectures, reduce costs associated with architectural design, and promote interoperability and standardization. The design process involved the following steps:

\begin{enumerate}
\item Literature review focusing specifically on autonomic architectures for smart buildings.
\item Literature review focusing on autonomic architectures developed in related computer science fields: 1 – big data analytics; 2 – cloud computing; 3 – networking.
\item A computational architecture concept based on MAPE-K was used as the starting point.
\item Conceptual layers were defined to ensure they were fully decoupled to avoid circular dependencies. Each layer in the architecture was defined to support one or more of the key autonomic system properties.
\item A mapping of AI sub-domains to each layer was developed.
\item A mapping of currently available technologies that could be used to implement each layer was provided to serve as an example of how to apply the reference architecture.
\item A control loop was added to support autonomic functionality.
\end{enumerate}

The B-SMART reference architecture was organized into three parts, or architectural views:

\begin{enumerate}
\item \textbf{The B-SMART High-Level Component Diagram}: This view describes the high-level logical components that comprise the reference architecture.
\item \textbf{The Functionality and Technology Layering}: This explains the static relationships between conceptual and technological layers and sub systems, and the responsibilities of each layer.
\item \textbf{The B-SMART Autonomic Control Loop}: This explains the dynamic relationships among the B-SMART technology layers and the relationship between the control loop and the SOCx processes.
\end{enumerate}

Sections ~\ref{comp-arch}, ~\ref{layering}, and ~\ref{ctrl-loop}, discuss these in greater depth.

\section{The B-SMART High-Level Component Architecture of Autonomic Smart Buildings} \label{comp-arch}

\begin{figure}
\begin{center}
\includegraphics[width=0.8\textwidth]{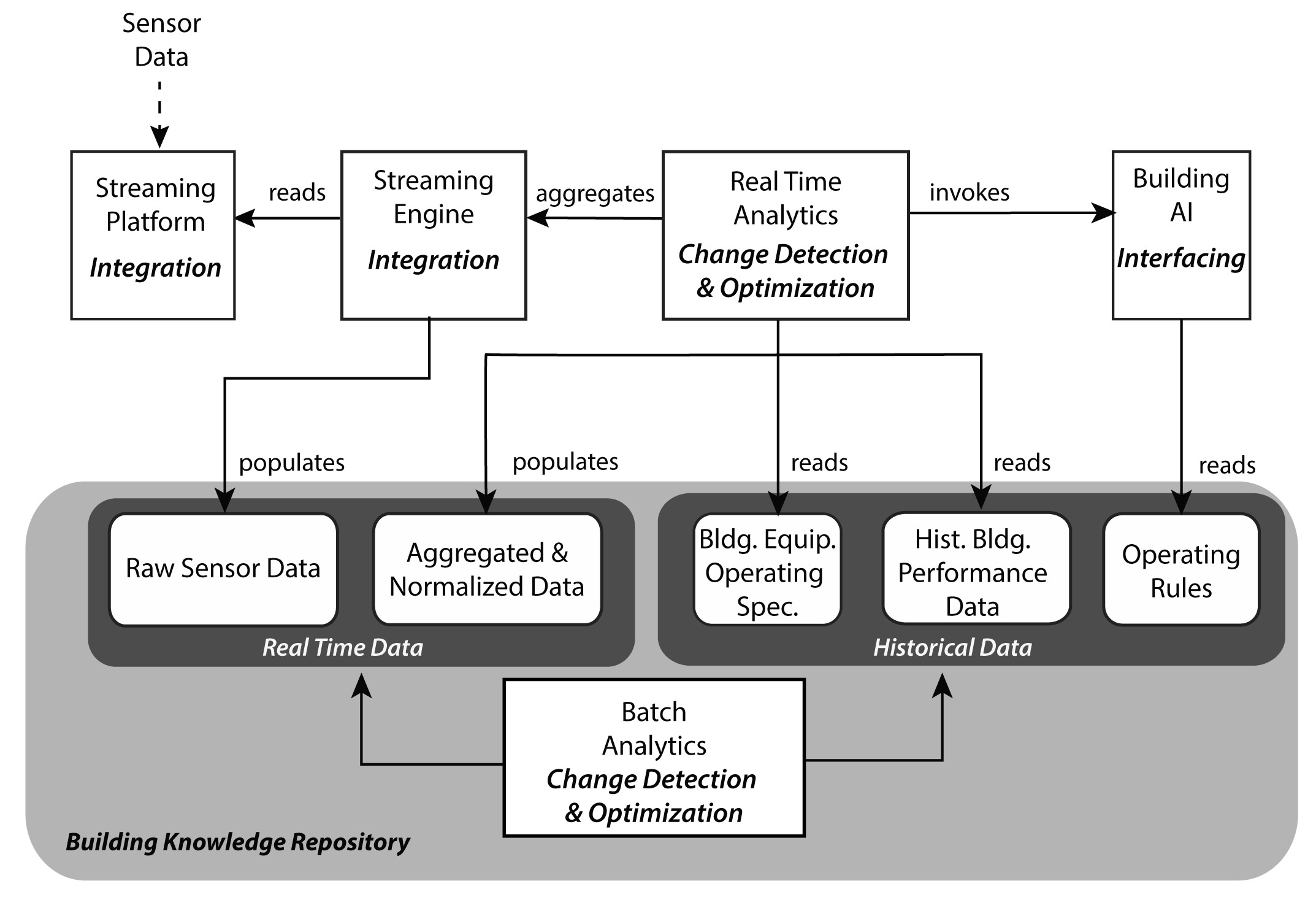}
\end{center}
\caption{\label{compdiag} The B-SMART high-level component diagram.}
\end{figure}

Figure ~\ref{compdiag} shows the high-level component architecture for autonomic smart buildings. This diagram also shows main categories of information that must be stored and accessed by the components, and the primary relationships between the components and the data. The arrows indicate architectural relationships among the high-level components rather then the flow of data. Corresponding functionality layers (discussed in sections below) from the B-SMART layered architecture view are shown in bold italics.

\subsection{Building Knowledge Repository (BKR)} \label{bkr}

As shown in Figure ~\ref{compdiag}, Knowledge is the central and essential component of the IBM MAPE-K autonomic manager reference architecture. The B SMART reference architecture builds on this concept and defines essential types of information and knowledge repository organization required by the smart building autonomic manager. The B-SMART reference architecture does not mandate any specific implementation technologies, deployment physical form factor such asan on-premises appliance for example, or physical topology (distributed vs. centralized) for the BKR.

The BKR must be able to store the following categories of information:

\begin{enumerate}
\item Description and classification types of all key building systems capable of generating data.
\item Manufacturer’s operating characteristics for sensors and actuators, and for the building systems.
\item Initial statistical baselines for all the smart building systems.
\item Historical information characterizing the performance of the smart building systems.
\item Real-time information characterizing current performance of the smart building’s systems.
\end{enumerate}

The rationale for requiring these information types is discussed in depth in sections focusing on Functionality and Technology Layering and the Autonomic Control Loop.

The BKR should be subdivided into two zones which manage the building operating data (see Figure ~\ref{compdiag}):

\begin{enumerate}
\item \textbf{Real Time Data Zone:} This zone contains data describing the current state of the smart building, and a small subset of data describing the building performance in the very recent (e.g. last 24 hours) past. These data are relatively volatile multi-variate time-series, that can include both raw data, results of real time data transformation, results of on-the-fly data aggregation, and near-real-time analytics. Performance characteristics of this zone must support high rates of inserts and updates as well as near-real-time operations on small volumes of data. The B-SMART reference architecture does not mandate any specific time segment duration that needs to be stored in the Real Time Zone, leaving this for individual smart building implementations to determine. We recommend, however, that the Real Time Zone maintain at least data collected over the previous 24-hour period, in addition to data describing the current state of the building. This is because the day-night cycle results in periodicity within smart building sensor data, that needs to accounted for as part of real-time analytic analysis. The degree of availability and fault tolerance which the Real Time Zone supports should be determined by individual smart building implementations based on each client’s ability to absorb the extra costs associated with sub-optimal energy performance versus the cost of supporting these qualities of service.
\item \textbf{Historical Data Zone:} This zone contains data describing the technical specifications and the historical performance characteristics of the building. Data could be in the form of multi-variate time series containing raw or transformed data, relations, analytic reports summarizing the building performance, and relations containing building equipment specifications. The Historical Data Zone may also contain persistent storage for rules used by the smart building AI. Performance characteristics of this zone must support fast analytics on large volumes of data (relative to the Real Time zone) and batch updates. The B-SMART reference architecture does not mandate any specific duration for the time segment of data that must be stored in the Historical Time Zone, but we recommend that at least one year of historical data is retained to enable predictive analytics that need to account for seasonal trends in the multivariate time series data. The Historical Time Zone must implement high availability and disaster recovery protocols because loss of these data will require re-commissioning of the smart building (discussed in the Section ~\ref{supporting-socx}).
\end{enumerate}

B-SMART components, and the manner in which they leverage the BKR are discussed in the following sections.

\subsection{Streaming Platform} \label{streamingplatform}

Messages containing sensor data generated by the smart building are transmitted to the Streaming Platform component via the smart building network infrastructure (Fabric layer code described in more detail in the section ‘Fabric’ below). The primary function of the Streaming Platform is to ensure that messages generated by the smart building sensors and actuators are not lost, and the temporal order of their arrival is preserved. It must allow for secure access to sensor data by the Streaming Engine component discussed in the next section. The Streaming Platform should support multiple queues or topics to allow separation of messages emitted by different sensors. For the B-SMART architecture, we recommend that the Streaming Platform support once-and-once-only guaranteed message delivery. This recommendation stems from the fact that smart building sensors send only change-of-value data. The Real Time Analytics component described in Sub-section ~\ref{rt-analytics} will need to impute data in order to construct a complete view of the smart building state. Repeat sensor messages can trigger unnecessary data imputation and optimization, degrading performance. B-SMART does not mandate the use of any specific streaming implementation technology for this component. Currently popular technological choices would be Apache Kafka \citep{kafka2022}, or Apache ActiveMQ \citep{acitveMQ2022}.

\subsection{Streaming Engine} \label{streamingengine}

Data stored in the Streaming Platform are read in near-real-time by the Streaming Engine component. The B-SMART reference architecture does not require true real-time processing semantics and leaves the establishment of this requirement to the discretion of the implementing team for each smart building.

A message stored in the Streaming Platform component contains data generated by one of the IoT devices (sensors and actuators) that instrument the smart building. Most sensors and actuators generate new messages only when they sense a change in the condition that they are monitoring – for example a change of value in room temperature above a defined threshold. The role of the streaming engine component is to read the disparate data messages stored in the Streaming Platform component and transform them into an integrated and coherent stream of multi-variate time-series data that can be consumed and operated on by the Real Time Analytics component described in the section below. The Streaming Engine thus:

\begin{enumerate}
\item Reads the messages from the topics or queues in the Streaming Platform
\item Transforms message data into formats that can be aggregated to form a cohered multi-variate time series
\item Imputes data that is missing at any given time (messages from various sensors arrive at disparate times)
\item Aggregates sensor data to form a coherent multi-variate time-series that describes the state of the smart building at any given time \textit{t}
\item Persists the integrated multi-variate time-series data in the Real Time Data Zone of the BKR at a pre-configured interval.
\end{enumerate}

The B-SMART reference architecture also does not mandate any specific streaming technology to implement the Streaming Engine component. Some examples of currently popular technologies include Apache Spark \citep{spark2022} and Apache Storm \citep{storm2022}.

\subsection{Real Time Analytics (RT Analytics)} \label{rt-analytics}

The RT Analytics component can run either in-process with the Streaming Engine, or in parallel. It consumes the raw event data from the Streaming Engine, aggregates and normalizes these data, and applies the change detection and real time optimization algorithms. This component persists the aggregated and normalized data into the BKR real time data zone. It also reads building equipment operating specifications and historical building performance data from the historical data zone of the BKR to enable change detection and on-line optimization algorithms.
These algorithms are discussed in more depth in the Sub-section ~\ref{cdo} in Section ~\ref{layering}.

The RT Analytics component invokes the Building AI component when it detects faults. The Building AI component is implemented by the Interfacing layer (discussed in more detail in Section ~\ref{layering}) code and reads and executes the operating rules stored in the historical data zone of the BKR to guide interactions with external actors (e.g. other smart buildings, facility maintenance personnel, other smart city elements such as smart grids, etc.).

\subsection{Batch Analytics} \label{batch-analytics}

The architecture of the autonomic smart building does not need to be strictly real time. The Batch Analytics component runs asynchronously, to perform Extract, Transform and Load (ETL) operations on the data stored in the BKR. Data stored in the real time zone must be periodically extracted, processed, and moved to the historical zone of the BKR to ensure that the real-time zone performance does not degrade with increasing data volume. The B-SMART reference architecture does not mandate how much data, either by volume or time horizon, must be stored in the respective zones of the BKR, and leaves this up to the specific implementation to determine.

The Batch Analytics component also executes the machine learning pipelines that are required to support the operations of the RT Analytics component. These pipelines can involve automated training and evaluation of supervised machine learning algorithms, execution of clustering and anomaly detection algorithms, automated statistical report generation, and simulations designed to support reinforcement learning approaches.

The B-SMART reference architecture does not mandate any specific implementation technology for the Batch Analytics component. Popular currently available choices include Apache Spark \citep{spark2022} and Apache Hadoop \citep{hadoop2022}, among others.

\subsection{Building AI} \label{building-ai}

The building AI component is a specialized AI trained to interact with external human and non-human actors to handle situations that cannot be resolved autonomically. The Building AI component is activated either explicitly by external actors, such as building maintenance personnel and tenants, or by the RT Analytics component when it detects faults. The Building AI component runs as a separate process to the Streaming Engine and RT Analytics components to allow building monitoring and optimization activities to continue while interaction with external actors takes place. The architecture of this component is discussed in the section below, as part of the discussion focusing on the architecture of the Interfacing layer.

\section{Functionality and Technology Layering} \label{layering}

\begin{figure}
\begin{center}
\includegraphics[width=0.8\textwidth]{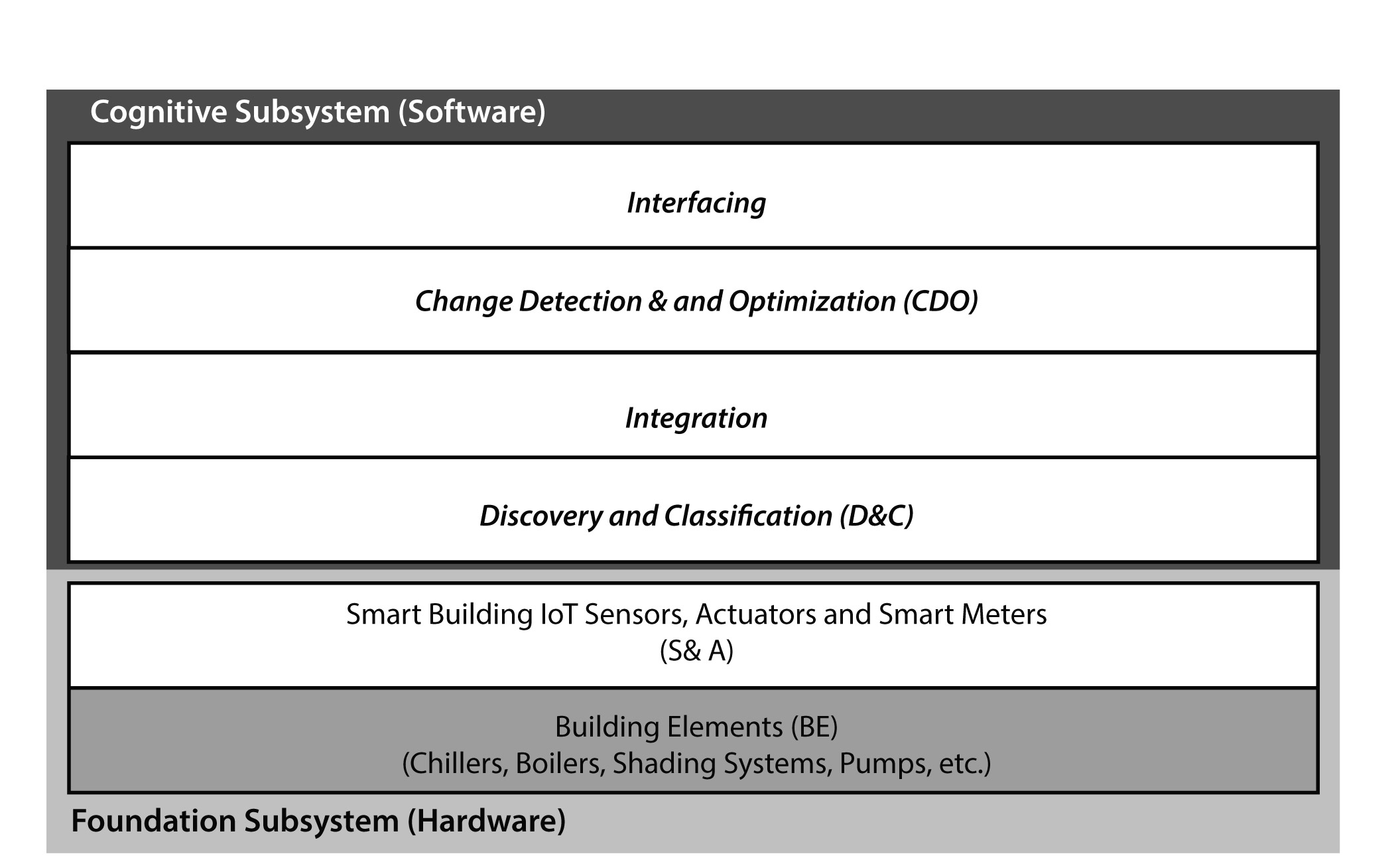}
\end{center}
\caption{\label{layers} The B-SMART layered architecture view.}
\end{figure}

In computer science and software engineering, the layered architectural pattern is used to establish the conceptual cascading dependency relationship between the sub-systems and components that comprise the architecture. The B-SMART layered architecture pattern is shown in Figure ~\ref{layers}. As with a building, lower layers of the architecture form the foundation. Each successive layer in the architecture builds its functionality using functionality provided by the lower layer. Functionality provided by each layer is discussed in the paragraphs below. Table ~\ref{layered-table} summarizes the autonomic properties that must be implemented by each layer of the architecture, and its key responsibilities.

\begin{xltabular}{\linewidth}{XXX}
\caption{The B-SMART Architectural Layers and Supported Autonomic Properties.} \label{layered-table} \\

\hline \multicolumn{1}{c}{Layer} & \multicolumn{1}{c}{Key Autonomic Properties} & \multicolumn{1}{c}{Responsibilities} \\ \hline 
\endfirsthead

\multicolumn{3}{c}%
{\tablename\ \thetable{} -- continued from previous page} \\
\hline \multicolumn{1}{c}{Layer} & \multicolumn{1}{c}{Key Autonomic Properties} & \multicolumn{1}{c}{Responsibilities} \\ \hline 
\endhead

\hline \multicolumn{3}{r}{{Continued on next page}} \\ \hline
\endfoot

\hline
\endlastfoot

Building Equipment&None&Must be capable of being instrumented with sensor and actuators.\\
Sensors and Actuators&Self-configuration \newline Self-description&Enable automation for the building equipment layer.\\
Fabric&Self-assembly&Ensures connectivity among the sensors and actuators and the cognitive
sub-system components. Sensors, actuators, and fabric networking elements form a self-organizing system.\\
Discovery and Classification&Self-learning&Searches the Fabric layer to discover and classify sources of data.\\
Integration&Self-organization&Organizes data sources discovered by the D\&C layer into more complex data flows. Incorporates legacy BIM/BAS.\\
Change Detection and Optimization&Self-optimization \newline Self-protection \newline Self-awareness \newline Self-regulation&Detects faults and concept drift in building systems. Mitigates faults and optimizes building performance in response to changes.\\
Interfacing&Self-description \newline Self-regulation&Interacts with human operators and systems external to the building.\\
\hline
\end{xltabular}

The layered view of B-SMART (Figure ~\ref{layers}) describes the static relationships between two autonomic smart building sub-systems – the Foundation Sub-system and the Cognitive Sub-system. In the following sections we provide more detailed descriptions of these sub-systems and their individual architectural layers.

In the foreseeable future, most smart buildings will be traditional buildings which have been converted to be smart. It is likely that in many cases this conversion will happen in stages, with each stage adding more and more ‘smart’ functionality to the building. Architectural layering of AI-related functionality encourages this process.

The Building Equipment (BE), the Sensors and Actuators (S\&A), and the Fabric layers form the Foundation Sub-System of the B-SMART reference architecture. The overall objective for this sub-system is to enable the operation of the Cognitive Sub-system – consisting of the Discovery and Classification (DC), Integration, Change Detection and Optimization (CDO), and Interfacing layers discussed in detail in section 4.2 – that implements the AI aspects of the autonomic smart
building.

Table ~\ref{layered-tech} summarizes the B-SMART architectural layers, the relevant sub-fields of AI research, and example technologies available today to help implement each layer. The B-SMART reference architecture does not promote any specific implementation technologies but encourages functional and technological layering for smart building implementations.

\begin{xltabular}{\linewidth}{XXX}
\caption{Mapping of AI research fields and the existing technology stack to the B-SMART Architectural Layers.} \label{layered-tech} \\

\hline \multicolumn{1}{c}{Layer} & \multicolumn{1}{c}{Relevant AI sub-fields} & \multicolumn{1}{c}{Existing Technologies} \\ \hline 
\endfirsthead

\multicolumn{3}{c}%
{\tablename\ \thetable{} -- continued from previous page} \\
\hline \multicolumn{1}{c}{Layer} & \multicolumn{1}{c}{Relevant AI sub-fields} & \multicolumn{1}{c}{Existing Technologies} \\ \hline 
\endhead

\hline \multicolumn{3}{r}{{Continued on next page}} \\ \hline
\endfoot

\hline
\endlastfoot
Building Equipment&N/A&N/A\\
Sensors and Actuators&N/A&Many vendors – e.g. Schneider Electric, ABB, Siemens, Honeywell, GE, others.\\
Fabric&Self-organizing networks&BACnet \citep{bacnet2022}, Modbus \citep{modbus2022}\\
Discovery and Classification&Unsupervised machine-learning, supervised machine learning, deep learning&TensorFlow \citep{tensorflow2022}, PyTorch \citep{pytorch2022}, Apache Spark \citep{spark2022}, scikit-learn \citep{scikit-learn}.\\
Integration&AI-assisted data mapping&There are gaps in this filed. Some currently available tools include: Talend \citep{talend2022}, and CloverDX \citep{cloverdx2022}. These require human interaction.\\
Change Detection and Optimization&Reinforcement learning, supervised machine learning, evolutionary computing&There are gaps in this field. Currently change detection algorithms need to be hand coded.\\
Interfacing&NLP, rules engines, rule induction&NLTK \citep{nltk2022}, TextBlob \citep{textblob2022} , Gensim \citep{gensim2022}, spaCy \citep{spacy2022}, polyglot \citep{polyglot2022}, scikit-learn \citep{scikit-learn}, others. Drools \citep{drools2022}, Easy Rules \citep{easyrules2022}, others.\\
\hline
\end{xltabular}

\subsection{The Foundation Sub-system} \label{found-sub-system}

\subsubsection{Elements (BE)} \label{be}

The BE layer groups all the key energy-related systems of the building and their constituent equipment, for example a cooling system consisting of a chiller, pumps, and terminal units or a
responsive shading system to minimize heat gains. This layer also includes additional energy producing and storage systems such as electrical panels, transformers, shading systems and capacitors. These systems do not necessarily need to be ‘smart’, but do need to be capable ofbeing instrumented with sensors and actuators to gather data. Most smart buildings will be conversions of traditional buildings that will, to differing extents, include existing sensors and/or sensor networks such as thermostats, building automation systems (BAS), etc. It is important to note that while the BE elements may be designed with integrated sensors and actuators, this is not mandated by the B-SMART reference architecture. Instead, B-SMART allows for monitoring and control technologies that are part of the S\&A layer to evolve separately because the computer and electronic hardware used to implement the sensors and actuators have historically evolved at a much faster rate than BE elements.

Because the BE layer is part of the Foundation sub-system of B-SMART, it should not be imbued with AI capabilities because local equipment optimization may miss the overall building optimization point. Instead, a whole-building approach is prescribed by B-SMART, reserving optimization to the higher levels where information is available from all relevant building systems, sub-systems, and interfaces.

\subsubsection{Sensors and Actuators (S\&A)} \label{sanda}

The S\&A layer communicates the inputs and outputs between the building elements and the B-SMART cognitive sub-system, transmitted via the fabric layer. Within the S\&A layer, the sensors serve as information sources for the higher layers in the reference architecture while the actuators enable actions to be performed on BE elements. While some actuators are physical, such as motorized valve actuators, others may be virtual, for example the signal to change the supply water temperature set point for the building’s boiler or chiller. Note that while a sensor or actuator may be integral to a BE element, the B-SMART architecture treats it as different conceptual entity to allow it to be separately configured and upgraded. This maximizes the system flexibility, allowing S\&A and BE technologies to evolve at different rates. S\&A devices could then be upgraded if necessary multiple times during the building life cycle. Table ~\ref{layered-tech} lists the currently available technologies that can be used to implement this layer.

\subsubsection{Fabric} \label{fabric}

The Fabric layer connects the sensors and actuators to each other and to the functionality in the higher Discovery and Classification (D\&C) layer. The Fabric layer is predominantly a hardware layer, with software-defined aspects, that consists of specialized controllers and networking elements such as WiFi routers, network switches, network wires, telephone cables, the required power supplies, and data storage needed to establish secure and reliable communications and support the functionality of the D\&C layer. The currently used BACnet \citep{bacnet2022} or Modbus \citep{modbus2022} protocols conceptually fits here as well. To support the cognitive function, however, these legacy systems are insufficient. The Fabric layer should ultimately be an autonomic networking layer – capable of supporting the self-creation, self-description, self-configuration, and self-management autonomic properties. Architectures summarized in \cite{movahedi2012} provide examples of how this can be accomplished.

\subsection{Cognitive Sub-system} \label{cognitive}

The Cognitive sub-system implements the main autonomic control loop of the building and supports the SOCx process. The overview of the functionality of each layer in the SOCx sub-system is given in the paragraphs below.

\subsubsection{Discovery and Classification (D\&C)} \label{dandc}

The role of the D\&C layer is to support the self-learning key autonomic system property by automatically finding and classifying different types of sensors, actuators, and other IoT devices available in the building. With the data feeds from the sensors and actuators on the S\&A layer, devices and their points can be classified by type using machine learning algorithms \citep{elmokhtari2021}. The D\&C layer relies on the data feeds exposed by the Fabric layer to be able to discover and connect to the different devices in the building. The D\&C layer updates the BKR with information describing the available data feeds and corresponding classifications, thus making it available to the Integration layer, described below.

\subsubsection{Integration} \label{integration}

The next layer in the B-SMART architecture is the Integration layer, which supports the self-organization autonomic property. This layer takes as input the raw, but identified and classified, data feeds from the D\&C layer and transforms them so that they can be effectively used by the higher cognitive layers. Data transformation operations can include mapping and transformation of data to a canonical format, normalization, imputation, filtering, anonymization, and security and confidentiality-related operations. Integration layer code also includes middleware used to implement the Streaming Platform and the Streaming Engine components (shown in Figure ~\ref{compdiag}).

\subsubsection{Change Detection and Optimization (CDO)} \label{cdo}

The CDO layer uses data stored in the BKR Real Time and Historical Data zones (see Figure ~\ref{compdiag}) to continually find the optimal operating parameters for the smart building. The CDO layer is also responsible for detecting changes in the operating characteristics of the smart building. Change detection is a very important stage in autonomic processing. Several researchers working on autonomic systems in various domains noted that continuous optimization incurs an overhead and reduces the eventual benefit (e.g. \cite{genkin2016}). To maximize the overall benefit, the frequency of parameter space searches must be optimized. This can be accomplished by searching the parameter space only when change is detected.

Changes can be broadly classified into two types: \textit
{faults} and \textit{concept drift}. Change detection can be implemented using a variety of techniques including, but not limited to, Bayesian statistical
methods, clustering algorithms, or supervised machine learning methods. These can be used to identify faults and concept drift affecting the building’s HVAC systems and other peripheral systems such as solar panels, shading systems, elevators, security, communications, etc. Computer vision techniques can be used to detect changes in building occupancy levels, broken pipes and other mechanical failures affecting the building systems, or other warning signs (e.g. flooding, fire, and smoke). CDO layer code is packaged and deployed as part of the RT Analytics and Batch Analytics components.

Once changes are detected, they can trigger one of three types of optimization response: a parameter search without engaging the Interfacing layer, a parameter space search engaging the Interfacing layer, or the direct engagement of the Interfacing layer without a parameter space search.

\subsubsection{Interfacing} \label{interfacing}

\begin{figure}
\begin{center}
\includegraphics[width=0.8\textwidth]{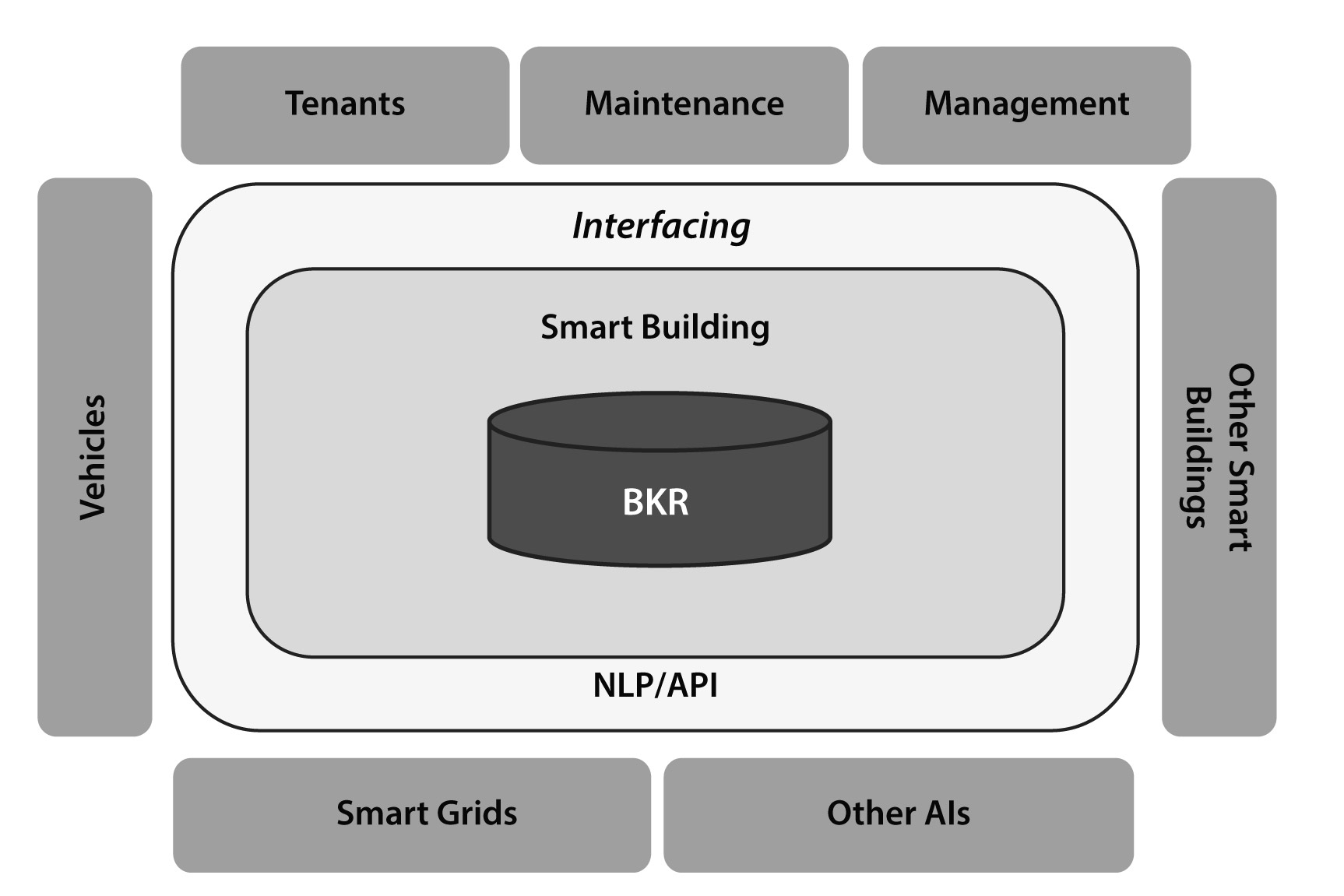}
\end{center}
\caption{\label{sysctxt} System context diagram for an autonomic smart building and the role of the Interfacing layer.}
\end{figure}

Serving as the top layer of the B-SMART, the Interfacing layer receives input from the CDO layer. When changes are detected in the operating characteristics of the smart building, this later interfaces with the appropriate external actor(s) to respond. These actions typically involve initiating workflows with humans in the loop, for example to replace faulty equipment. Figure ~\ref{sysctxt} shows the system context diagram for an autonomic smart building, and the role of this layer.

As implied by this figure, the Interfacing layer hosts a specialized AI that provides the cognitive functionality necessary to maintain and operate the building and interact with several human and non-human external actors. These include:

\begin{enumerate}
\item \textbf{Building tenants:} The human inhabitants of the smart building need to be able to interact with the building to adapt the building environment to suit their comfort level. Currently a common interaction involves adjusting the temperature setting of the smart meter or filling in a Web-based form indicating the comfort level or lack there of. Future interaction techniques should focus on NLP and video as primary interaction channels between the tenant and the smart building AI. The interactions between the tenants and the smart building could be initiated by either party, but most commonly it will be the tenants who will contact the smart building AI to address a particular concern.
\item \textbf{Building maintenance and management personnel:} This is another very important category of human actors that the smart building needs to interact with. These interactions could be focused on routine maintenance or as a response to detected faults. For example, if the smart CDO layer detects that the sensors responsible for controlling the building chiller are reporting erroneous values, the Interfacing layer will contact the appropriate building maintenance technician. To determine which technician to contact, and how to contact him or her, the Interfacing layer may engage a rules engine. While today most interactions will involve contacting the maintenance technician via e-mail, or text message, future directions should encourage NLP voice interactions between the smart building AI and these human actors.
\item \textbf{Human-driven vehicles:} Vehicles frequently need to interact with smart buildings. One example of such an interaction could involve a human driver contacting the smart building to enquire about the availability of parking spaces and charging stations for
electric vehicles. The B-SMART architecture encourages the use of voice and video channels and NLP techniques to accomplish these types of interactions.
\item \textbf{Smart grids:} Vehicles frequently need to interact with smart buildings. One example of such an interaction could involve a human driver contacting the smart building to inquire about the availability of parking spaces and charging stations for
electric vehicles. The B-SMART architecture encourages the use of voice and video channels and NLP techniques to accomplish these types of interactions.
\item \textbf{Other smart buildings:} Other smart buildings may have indirect relationships with the target building, for example through connection to the same smart grid, or direct relationships, for example through shared ownership, occupants, or building systems.
\item \textbf{Autonomous driving vehicles:} As with human-driven vehicles autonomous driving vehicles will need to interact with smart buildings. B-SMART architecture encourages the use NLP, rather than APIs, as the universal communication interface for both human-driven and autonomous vehicles. The main advantage of this approach is that NLP-based communication can be readily understood and audited by humans.
\item \textbf{Other AIs (responsible for complimentary domains):} The smart building AI will likely need to communicate with other AIs, such as Google’s Google or Apple’s Siri, to perform its functions or accommodate queries from its residents. One example could involve an inquiry by a building resident about upcoming weather conditions in the immediate vicinity of the smart building. This query, after pre-processing to augment it with the precise location of the building, could be passed on to another AI that specializes in this type of query. B-SMART encourages the use of NLP techniques for these types of interactions to facilitate auditing by humans.
\end{enumerate}

\begin{figure}
\begin{center}
\includegraphics[width=0.8\textwidth]{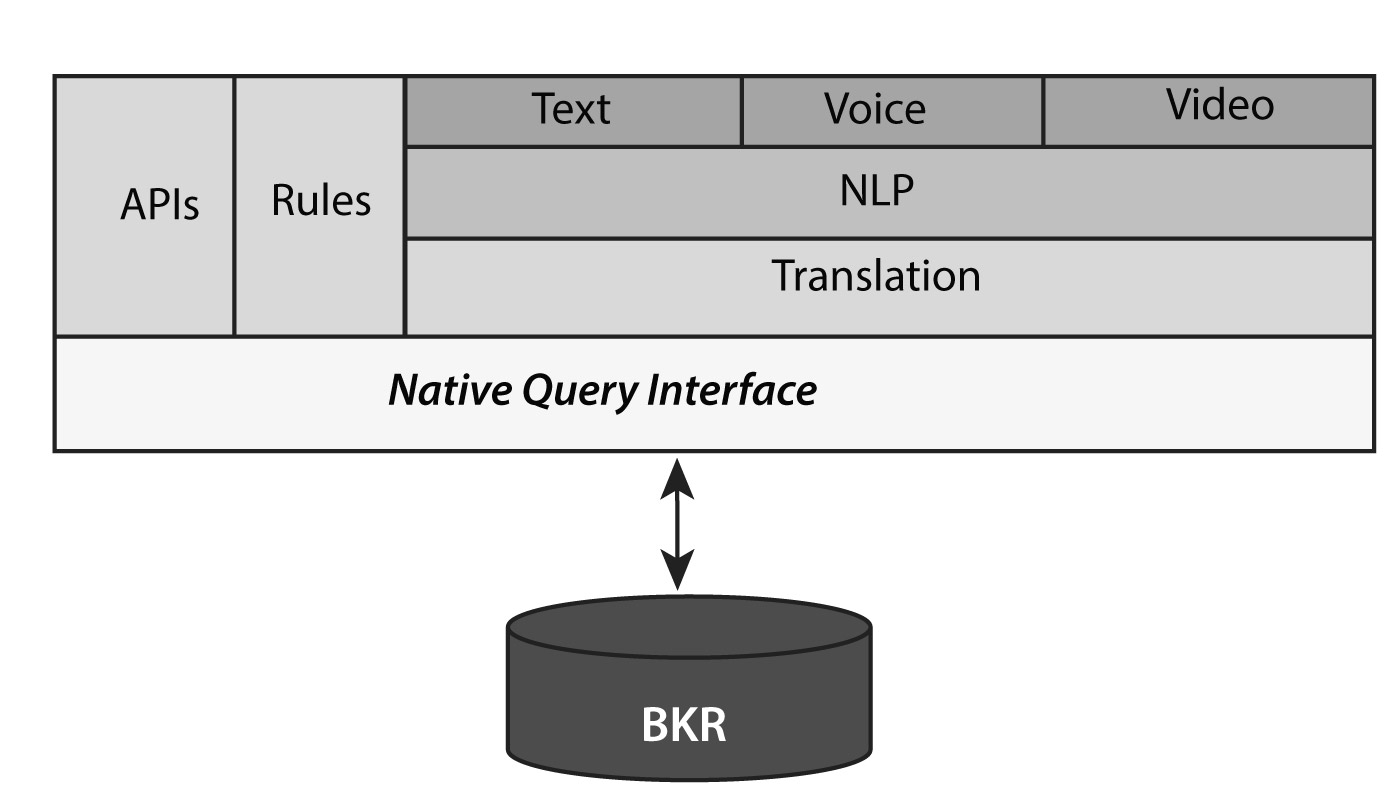}
\end{center}
\caption{\label{interfacinglayers} Interfacing layer functionality and technology sub-layers.}
\end{figure}

The Interfacing layer itself should be implemented using the layered architectural pattern shown in Figure ~\ref{interfacinglayers}. The Interfacing layer leverages the information stored in the BKR to establish the current state of the smart building and to communicate relevant information to human and non-human autonomic actors. The NLP sub-layer supports Text, Voice, and Video communication with human actors. The B-SMART reference architecture strongly encourages the use of NLP-based communications because they can be easily understood and audited by human actors with little to no IT and computer science knowledge. As discussed earlier in this work, the need to hire or acquire IT skills is a major inhibitor in smart building adoption. The Translation sub-layer converts verbal and text natural language queries into queries that can be understood by the Native Query Interface sub-layer that is used by the BKR implementation. The B-SMART reference architecture does not mandate any specific implementation for the Native Query Interface sub-layer. Each commercial or open-source technology will likely provide their own database-specific implementation of this layer. The B-SMART reference architecture encourages the development of new technologies to implement the NLP and Translation sub-layers.

To fully implement the self-description property, the smart building must be able to query the BKR and return the relevant information to the correct external actor. Interactions between the smart building and human actors could include financial reporting, technical reporting, proactive maintenance requests, interactions with the building occupants regarding their comfort level, and reactive maintenance requests. These interactions could be implemented using either exposed APIs or via using NLP.

To implement self-healing, configurable rules would determine which actions need to be performed for a given failure scenario, and in which sequence. For example, if a power failure is detected, the smart building could automatically start the back-up power generator. The building could also send notifications of the event to human actors responsible for building maintenance. Once the building detects that the main power is back on, it would stop the back-up power generator and switch to the main power supply. There are many other failure scenarios, some of which may involve self-healing, while other would require intervention by external human or autonomic actors.

\section{The B-SMART Autonomic Control Loop} \label{ctrl-loop}

\begin{figure}
\begin{center}
\includegraphics[width=0.8\textwidth]{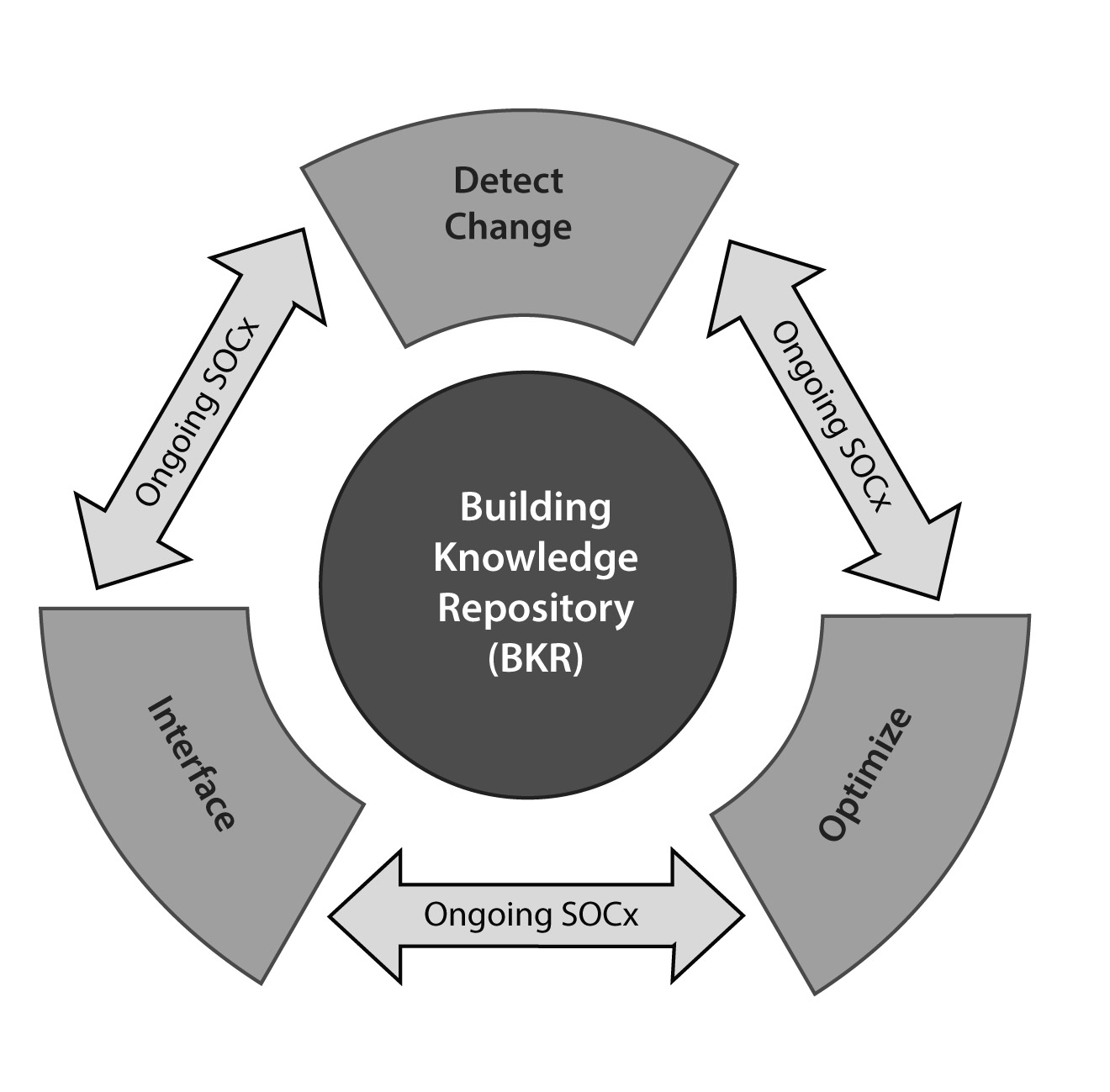}
\end{center}
\caption{\label{ctrlloopdiag} The B-SMART autonomic cycle.}
\end{figure}

In this section, we present the Autonomic Control Loop of the B-SMART reference architecture. The autonomic cycle view of our reference architecture, which describes the dynamic relationships between the smart building sub-systems, is shown in Figure ~\ref{ctrlloopdiag}. Table ~\ref{ctrlstages} summarizes the main stages of the autonomic control loop, the B-SMART architectural layers involved in each stage, the most relevant AI sub-fields, and the currently available technologies that can be used to implement each stage.

\FloatBarrier
\begin{table}[h]
\caption{\label{ctrlstages} Mapping of the most relevant AI research fields and the existing technology stack to the B-SMART autonomic control loop stages.}
\begin{center}
\begin{tabularx}{\textwidth}{XXXX}
\hline
Stage&Involved Layers&Relevant AI Sub-fields&Applicable AI Technologies\\
\hline
Monitor&S\&A, Fabric, CDO&Computer vision&OpenCV \citep{opencv2022}\\
Detect Change&CDO&Change detection algorithms, unsupervised machine learning, supervised machine learning&Apache Spark \citep{spark2022}, scikit-learn \citep{scikit-learn}, SciPy \citep{scipy2022}\\
Optimize&CDO&Reinforcement learning, supervised machine learning, evolutionary computing&Open-AI \citep{openai2022}, Keras-RL \citep{kerasrl2022}\\
Interface&Interfacing&NLP, Rules engines&TensorFlow \citep{tensorflow2022}, PyTorch \citep{pytorch2022}\\
\hline
\end{tabularx}
\end{center}
\end{table}
\FloatBarrier

\subsection{Monitoring, Optimizing, and Interfacing} \label{mon-opt-int}

The autonomic smart building will spend most of its time simply monitoring the building state (Monitor stage) rather than performing actions. The actual monitoring of the building state should be handled entirely by the CDO layer code packaged in the RT Analytics component, which consumes data generated by the S\&A layer, transported by the Fabric layer, and/or processed by the Integration layer.

The CDO layer compares the data in the current observation window with historical data stored in the BKR Historical Data zone. If change is detected, it is first classified. If the change is classified as a \textit{fault}, then the CDO layer passes all the relevant information to the Interfacing layer and continues processing. The Interfacing layer further analyzes the fault and determines the correct set of actions that need to be executed to address the fault. If the change is classified as \textit{concept drift} the CDO layer performs a parameter space search (Optimize stage in Fig. ~\ref{ctrlloopdiag}) to
determine the optimal parameters for the new set of conditions. The newly discovered optimal set of parameters is then stored in the BKR.

The CDO layer code continuously monitors the state of the BE systems, detects changes and updates BKR as required. Interactions between the CDO layer and the Interfacing layer should be asynchronous, allowing for continuous monitoring and optimization of different smart building systems even as some of the systems may be taken offline due to faults and/or scheduled maintenance.

The B-SMART reference architecture does mandate real-time or near-real-time semantics for the Monitor and Detect Change stages. The architecture could be implemented using batch semantics as well, with important processing, such as machine learning algorithms training, happening during operational maintenance windows. B-SMART does encourage asynchronous interaction semantics between the CDO and the Interfacing layers because this leads to a more resilient and fault tolerant overall architecture.

Interaction with external human and non-human actors should be handled entirely by the Interfacing layer. Once the CDO layer identifies faults that need to be handled by the Interfacing layer it passes context information to the latter. The Interfacing layer then establishes the most appropriate sequence of actions that need to be executed in response to this fault and executes them (\textit{Interface stage} in Fig. ~\ref{ctrlloopdiag}, \textit{Interfacing state} in Fig. ~\ref{statediag}).

These actions should be performed asynchronously and in parallel. Consider the following scenario. At time t the CDO layer may detect that the building chiller is functioning outside its documented operating parameters. The CDO layer passes the context describing this situation to the Interfacing layer and, in parallel, performs a parameter space search on the chiller to establish the new optimal schedule for temperature set-points to use. At time \textit{t+1} the CDO layer determines that a power failure has occurred. The CDO layer passes this observation to the Interfacing layer to handle. In response to these events the Interfacing layer:

\begin{enumerate}
\item At time \textit{t}, contacts building maintenance personnel and/or vendor responsible for the chiller maintenance and interacts with them to schedule a maintenance visit.
\item At time \textit{t+1}, starts the building backup-power generator.
\item Contacts building maintenance personnel responsible for power maintenance, contacts the smart grid and/or smart city points responsible for information that describes the state of the power grid.
\item Contacts the building tenants and management personnel to notify them of the power
failure and provide an update on when normal power service is expected to be restored.
\end{enumerate}

While these interactions are happening the CDO layer continues monitoring the state of the building systems. Because the power failure at time \textit{t+1} is more serious it will likely be handled and fixed by the maintenance personnel before the chiller issue determined at time \textit{t}. While waiting for maintenance the chiller will continue to operate using the newly established optimal schedule of temperature set-points.

Once the maintenance personnel perform chiller maintenance, they will confirm this to the Interfacing layer. The Interfacing layer will notify the CDO layer (asynchronously) that a new parameter space search needs to be performed on the building chiller. The CDO layer performs the parameter space search on the chiller and stores the new optimal parameters in the BKR. These types of interactions between the Interfacing and the CDO layers f rm the on-going phase of the SOCx process.

\subsection{Smart Building State Transitions} \label{state-transitions}

\begin{figure}
\begin{center}
\includegraphics[width=0.8\textwidth]{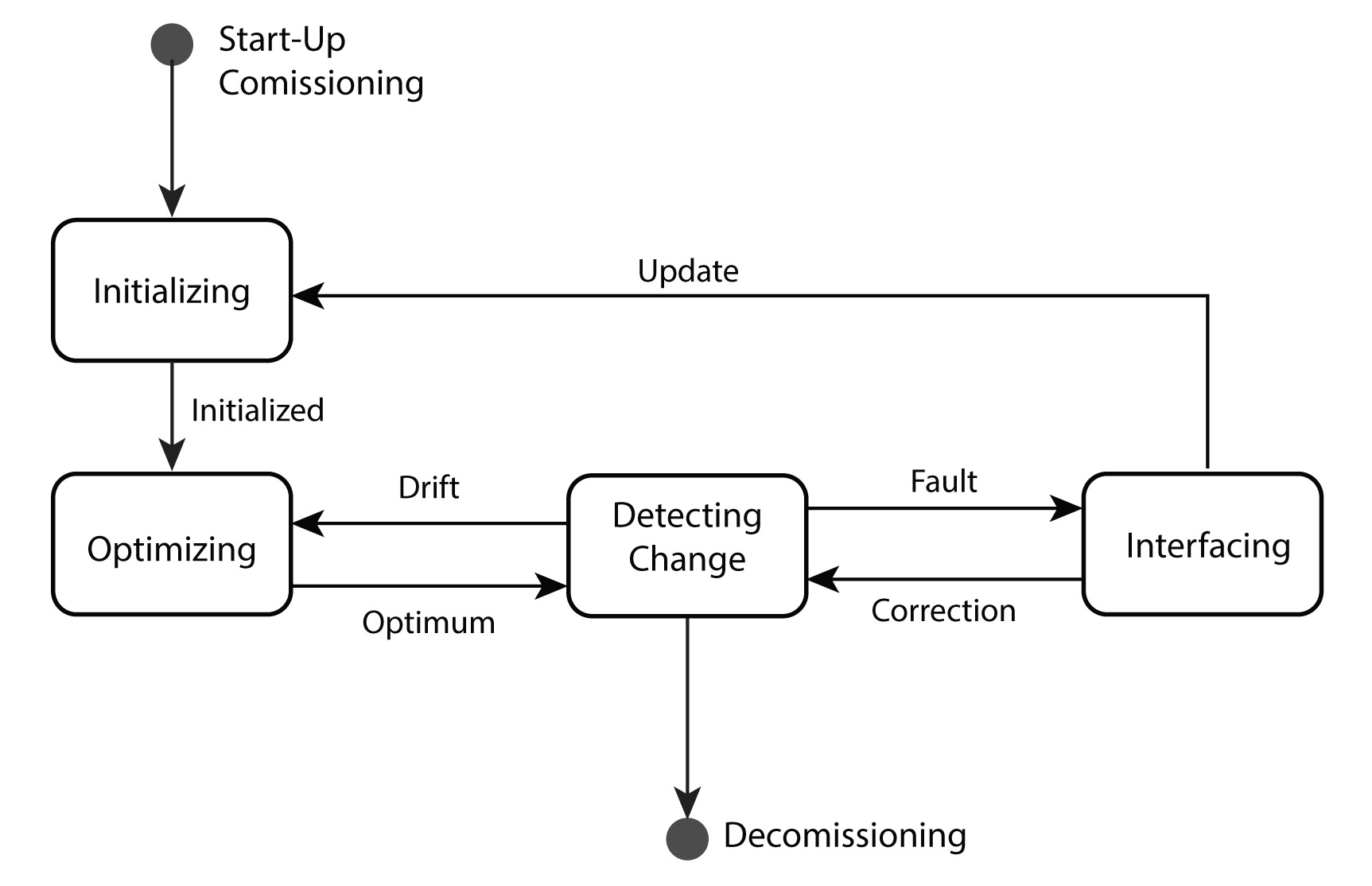}
\end{center}
\caption{\label{statediag} The B-SMART building state transitions.}
\end{figure}

Figure ~\ref{statediag} presents the state transition diagram for an autonomic smart building. While the start-up commissioning SOCx process is on-going the autonomic smart building remains in the‘Initializing’ state. Once the start-up SOCx process deploys the BKR and populates it with initial operating parameters the building transitions into the ‘Optimizing’ state, because, at this point in time, the optimal operating parameters for the building systems have not yet been established. Once the optimal operating parameters have been found the building transitions into the ‘Detecting Change’ state.

The smart building remains in this state until the CDO layer code detects one of two change events. If the change event is classified as concept drift, the smart building transitions into the ‘Optimizing’ state, during which the parameter space search is performed. Once the new optimum is found the smart building transitions back into the ‘Detecting Change’ state. If the
change is classified as a fault, the smart building transitions into ‘Interfacing’ state. While in this state the Interfacing layer code interacts with external actors to resolve the fault condition. Once the fault is resolved, if repairs do not involve changing, adding, or upgrading BE systems, the smart building transitions back into the ‘Detecting Change’ state.

If the building systems are changed, added, or upgraded, the maintenance personnel will send a message to the Interfacing layer to put the building into the ‘Interfacing’ state. During the upgrade procedure the building will transition into the ‘Initializing’ state. Once the new initial operation characteristics have been entered into the BKR, the building transitions into the ‘Optimizing’ state. This flow is exactly the same as for the start-up commissioning process described above.

\subsubsection{Reacting to Changes in the Building Equipment} \label{equipment-changes}

Conversion from a traditional building to a smart building will likely be performed iteratively. As new sensors and BE systems are added to the building, external actors responsible for installing them notify the Interfacing layer. Sensor data feeds are automatically discovered by the
Fabric layer, classified by the D\&C layer, and integrated into the data stream by the Integration layer.

The Interfacing layer communicates with the external actors to populate the BKR with information describing the newly added systems and sensors and rules describing how to handle failures affecting these systems.

The CDO layer automatically initiates parameter space search on any newly discovered systems. Once optimal operating parameters are established, they are stored in the BKR. The CDO layer adds the newly installed systems to the monitoring and change detection activities. These interactions comprise the SOCx on-going commissioning process.

\section{Integrating with Legacy Building Automation Systems (BAS)} \label{basintegration}

Many traditional buildings are equipped with legacy BAS. These systems may implement some of the capabilities described in B-SMART but will not have the same degree of autonomic functionality and functional and technological decoupling.

Legacy BAS are not currently capable of automatic discovery and classification of the different sensors and actuators but rely on manual configuration, the D\&C functionality must be provided separately. In those situations when BAS expose APIs that allow it to send messages with payload describing the state of systems managed by the BAS to the Streaming Platform they can be integrated with the other smart building components.

The Integration layer code packaged in the Steaming Engine can then incorporate these data into the multi-variate time-series that form the smart building data stream. These data can then be captured in the Real Time and Historical Data Zones of the BKR. These data can then be operated on by the CDO layer code packaged in the RT Analytics component (see Figure ~\ref{compdiag}).

Legacy BAS systems also typically provide dashboards accessible over the Web. This allows for an on-the-glass integration strategy in addition, or as an alternative to, message-level integration. This integration strategy will require the Interfacing layer code packaged in the Building AI component (see Figure ~\ref{compdiag}) to display and/or describe using NLP contents of the dashboard elements exposed by the legacy BAS.

The B-SMART preferred integration strategy is message-level integration, on the middle tier, via message-oriented middleware. This successful integration approach was described, for example, in a case study by El Mokhtari et al. (El Mokhtari, et al., 2022). Alternate integration strategies involving the database and presentation tier are likely to be complicated by the fact that legacy systems need to adhere to more limited data models and user interfaces than those implemented using the latest AI technologies.

\section{Supporting the SOCx Processes} \label{supporting-socx}

There are two SOCx processes to consider: 

\begin{enumerate}
\item Start-up Commissioning
\item On-going commissioning.
\end{enumerate}

Autonomic smart buildings must support both processes. Sections below discuss how this can be accomplished in more detail.

\subsection{Start-up Commissioning} \label{start-up}

Start-up commissioning of smart buildings (as different from and in addition to activities performed during the commissioning of traditional buildings) must include:

\begin{enumerate}
\item Identifying all of the sensors and actuators used by the BE layer systems.
\item Classifying them by type.
\item Ensuring that the data being generated by the sensors has values within the standard operating parameters established by the manufacturers of both the sensor and the BE system that it instruments.
\item Capturing and storing the statistical baselines for each piece of equipment.
\item Creating and populating the BKR with above data.
\end{enumerate}

Start-up commissioning involves leveraging the functionality provided by the D\&C layer and the Integration layer code to establish the data streams that will continuously update the BKR. The autonomic control loop starts once the start-up commission process successfully completes. Monitor functionality of the CDO layer is used to analyze the incoming data. Change detection functionality of the same layer is used to identify situations that require an action to be performed by the autonomic smart building manager.

The Integration layer includes and supports the Start-Up Commissioning process. Another way to look at this relationship is to consider that one of the key goals of the start-up commissioning process for a smart building must include achieving tight integration of all sensors and actuators, and legacy BIM/BMS, with layers implementing higher-level autonomic functionality. The start-up commissioning process bootstraps the main autonomic control loop of the smart building by populating the BKR.

The Interfacing layer coordinates both the start-up commissioning SOCx stage, and the on-going SOCx commissioning activities. While the ultimate goal is to fully automate these activities, it is understood that at least in the near-term future most will require interaction with human actors.

\subsection{On-going Commissioning (OCx)} \label{ocx}

Throughout the building lifecycle the CDO layer and Interfacing layer code update data stored int the BKR. Building state transitions shown in Figure ~\ref{statediag} comprise the on-going commissioning process of the SOCx. Whenever the CDO layer code detects change and triggers an optimization of the building operating parameters, this action in effect represents automated re-commissioning of some of the smart buildings systems. Once the new optimum is found it is recorded in the BKR, creating a new baseline for the smart building. Transition from the Optimize stage to the Detect Change stage also forms a part of the on-going commissioning process.

Once the CDO layer code identifies a fault and invokes the Interfacing layer to engage external actors, such as the building maintenance staff, it has in effect triggered re-commissioning of some of the buildings systems and forms a part of the on-going commissioning process. The on- going SOCx process can be generalized to involve the following steps:

\begin{enumerate}
\item Change detection. If the building is in a steady state, then no re-commissioning of systems is needed.
\item Repair or replacement. If the change is a fault, then external actors need to repair or replace some of the buildings systems. If the change is not a fault, then this step is skipped.
\item Optimization. Performance of newly repaired or replaced components must be automatically optimized.
\item Update. Newly established optimal operating characteristics need to be stored in the BKR and become the new baseline for the CDO layer code.
\end{enumerate}

The B-SMART autonomic reference architecture is thus designed to natively support both the start-up commissioning and on-going commissioning processes for smart buildings.

\section{Applying B-SMART - A Case Study} \label{case-study}

As noted in the Section ~\ref{integration}, reference architectures are used to accelerate the software and systems design cycle. This section presents an example case study during which we applied the B-SMART reference architecture on an existing smart building to map out the research and development road map for future enhancements.

The Daphne Cockwell Complex (DCC) building of the Toronto Metropolitan University campus has been a showcase smart building and was used for our case study. Smart building research conducted on the DCC has been presented at many scientific conferences \citep{misic2021, mokhtari2022}. Although this building has been extensively instrumented and studied, the general consensus is that it's journey to becoming a truly autonomic smart building is far from complete. In sub-sections below we first present a fit-gap analysis of the DCC relative to our B-SMART reference architecture.

\subsection{Fit-Gap Analytis of DCC}

The fit-gap analysis of the current DCC architecture vs. B-SMART was performed. To accomplish this the following methodology was followed:

\begin{enumerate}
\item Compare the current DCC architecture to the B-SMART high-level component diagram (Figure ~\ref{compdiag}) to identify missing conceptual elements.
\item Compare current DCC interfacing capabilities to the B-SMART interfacing capabilities identified in the Section ~\ref{interfacing}, and shown in Figures ~\ref{sysctxt} and ~\ref{interfacinglayers} to identify missing interfacing capabilities.
\item Compare current DCC autonomic control capabilities to the B-SMART autonomic control capabilities identified in the Section ~\ref{ctrl-loop}, and shown in Figure ~\ref{ctrlloopdiag} to identify missing capabilities required to implement the building autonomic control loop.
\end{enumerate}

\begin{figure}
\begin{center}
\includegraphics[width=0.8\textwidth]{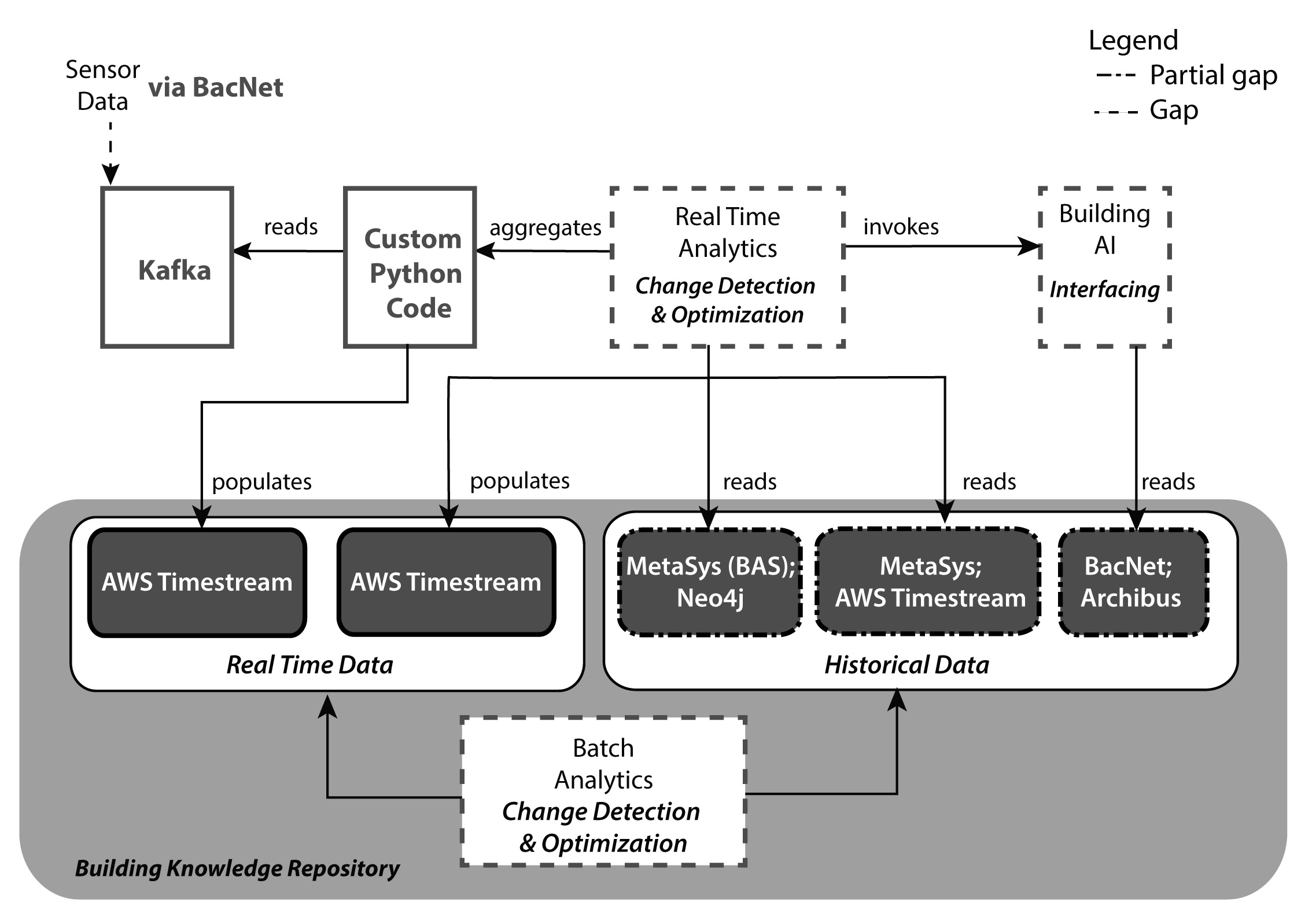}
\end{center}
\caption{\label{compdiag_appl} The DCC technical solution mapped onto the B-SMART high-level component diagram.}
\end{figure}

The Figure ~\ref{compdiag_appl} shows the DCC technical solution - technologies currently being used to implement the operating DCC systems - mapped onto the B-SMART high-level component diagram.

The DCC Streaming Platform component is implemented using the Apache Kafka messaging platform deployed in the Amazon Cloud. Custom Python scripts are used to route change-of-value messages from the DCC BAS system to topics on the Kafka broker.

The Streaming Engine component is implemented using custom Python code that picks up messages from Kafka topics, extracts payload contents, transforms the contents to the format required by the Amazon Timesteam database, and persists that data in that database.

The BKR is currently implemented using the Amazon Timestream database. This database is optimized for storing and working with time-series data. The database stores raw sensor change-of-value data. This database is used to store both real-time data, and to perform historical data analysis. Building equipment specifications and the relationships among them (ontology) are stored in the neo4j graph database. Although there is no dedicated rules engine, some of the building operating rules are stored in the DCC Archibus data base.

Notable gaps (see Figure ~\ref{compdiag_appl}) are:

\begin{itemize}
\item \textbf{Bulding AI}.
\item \textbf{RealTime Analytics}.
\item \textbf{Batch Analytics}.
\end{itemize}

Currently there is no Building AI that oversees, monitors, and coordinates the DCC systems, and is capable of interacting with other actors as described in the Sub-section ~\ref{building-ai}. Although some of the building systems are capable of semi-autonomous operation, they are not coordinated in any way. 

For example, the DCC is equipped with an adjustable shading system. The automatically operated shades adjust position based on a pre-programmed schedule. The building cooling and heating systems also adjust the chiller and boiler set points based on a pre-programmed schedule. These schedules, however, are pre-programmed separately by human maintenance personnel. The two systems are not aware of each other.

In many cases it may be possible to maintain the building comfort level by adjusting the shading system without changing the chiller or boiler set point, and this would likely result in additional power savings. This, however, is not currently possible due to a lack of an AI that is integrated with all key building systems, and capable of coordinating their operation.

The DCC IT architecture also currently lacks fully functional RT Analytics and Batch Analytics components. The RT Analytics component, as discussed in the Sub-section ~\ref{rt-analytics} must be capable of aggregating and normalizing data in real-time. To accomplish this it needs to be capable of imputing missing data on-the-fly, because the DCC BAS dispatches only change-of-value messages. This component needs to be capable of performing these operations, and then performing statistical computations, on incoming data and comparing the results with historical data aggregated and produced by the Batch Analytics component. While research projects are currently under way to address this gap, the results are not currently deployed in production. This is further discussed in the Sub-section ~\ref{tech-road-map}.

\begin{figure}
\begin{center}
\includegraphics[width=0.8\textwidth]{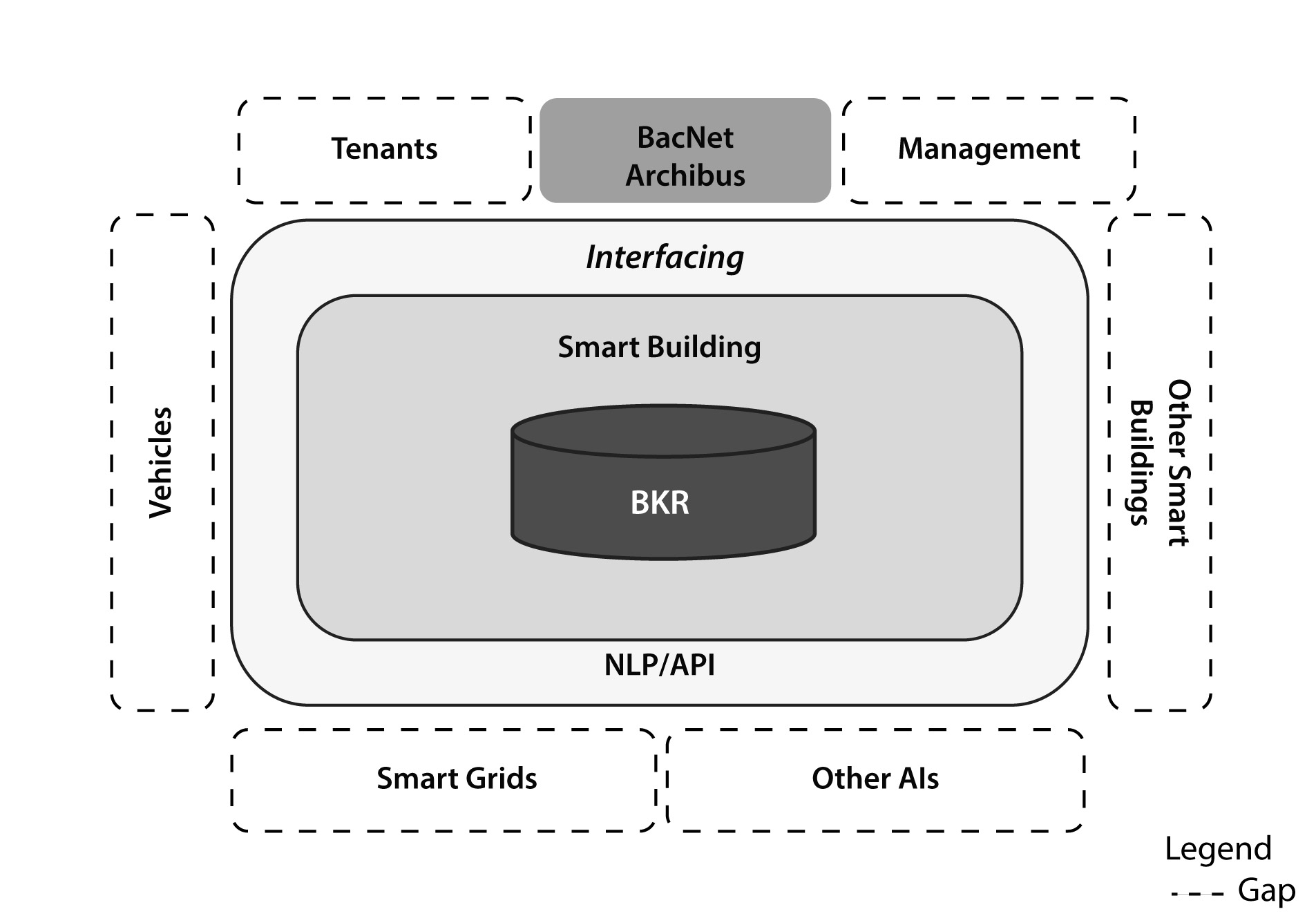}
\end{center}
\caption{\label{sysctxt_appl} The DCC technical solution mapped onto the B-SMART system context diagram to highlight interfacing gaps.}
\end{figure}

Figure ~\ref{sysctxt_appl} shows the results of fit-gap analysis of the DCC interfacing capabilities vs. the B-SMART system context diagram ~\ref{sysctxt} and ~\ref{interfacinglayers}. Currently the DCC is able to interface only with the building maintenance personnel via the DCC BAS Web-based interface that serves status information for the building systems as HTML pages.

The DCC is not currently capable of interfacing with the building tenants, management personnel, other smart buildings, other AIs, smart grids, or vehicles. The DCC BAS, which is the only system capable of interfacing with external actors, has no NLP capability.

\begin{figure}
\begin{center}
\includegraphics[width=0.8\textwidth]{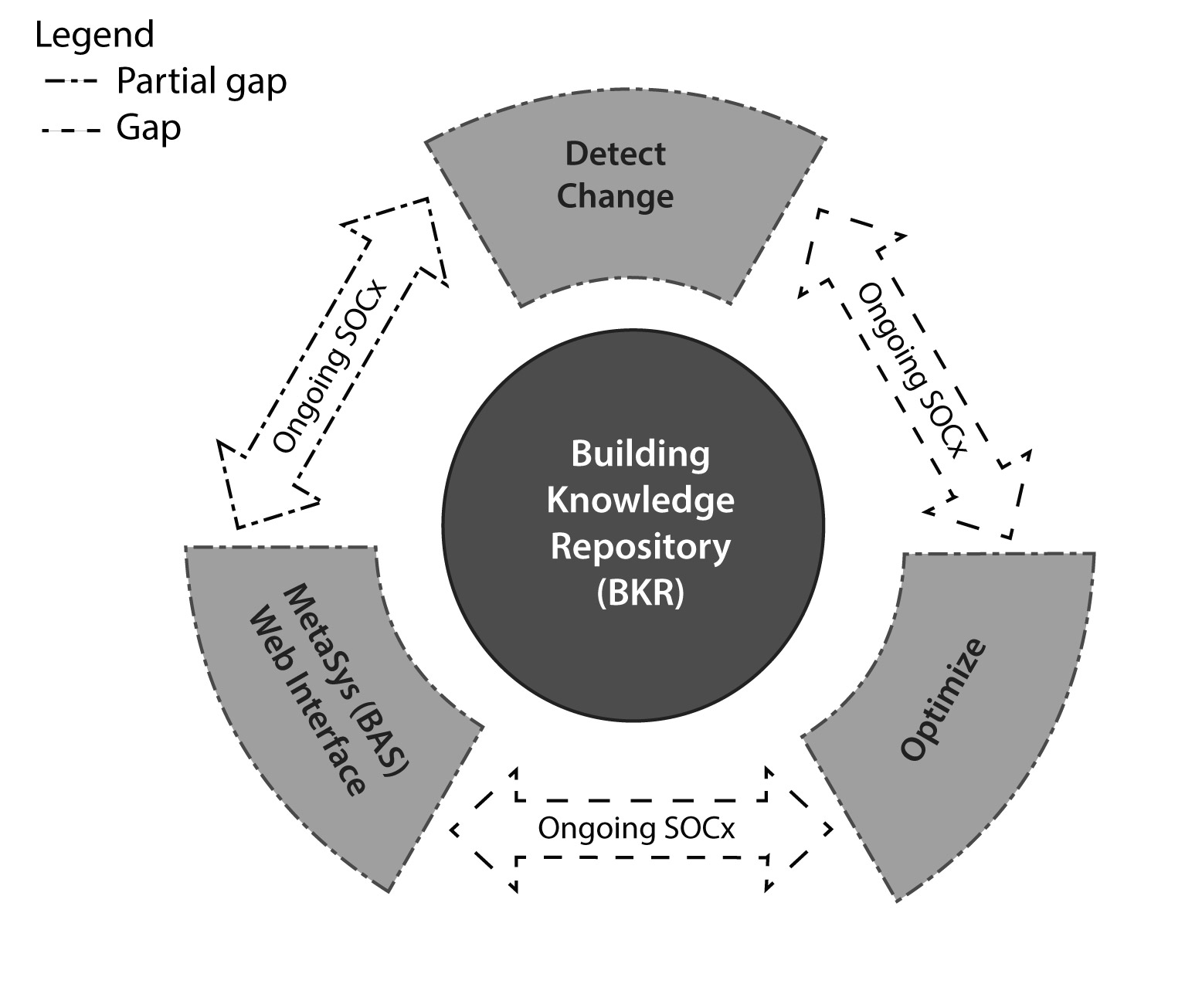}
\end{center}
\caption{\label{ctrlloopdiag_appl} The DCC technical solution mapped onto the B-SMART autonomic control loop diagram.}
\end{figure}

Figure ~\ref{ctrlloopdiag_appl} shows the results of fit-gap analysis vs. the B-SMART autonomic control loop shown in Figure ~\ref{ctrlloopdiag}. Currently, the Detect Change, Interface, and Optimize stages of the autonomic loop are only partially implemented. While there are research projects in progress to address the gap needed to fully implement the Detect Change stage, this effort needs to continue until it is deployed in production (see discussion in the sub-section below). 

Limited interfacing capability exists via the DCC BAS Web-based interface. This allows the DCC to transition from the Detect Change stage to the Interface stage and back. The DCC BAS is not currently capable of automatically transitioning from the Detect Change stage to the Optimize stage and back, and from the Optimize stage to the Interface stage and back. Currently these transitions would have to be completely manually-implemented  human-driven procedures.

Additional detail can be gleaned by performing fit-gap analysis on Figures ~\ref{interfacinglayers} and ~\ref{statediag}. Information obtained by doing so can be used during the detailed design stage to refine the newly proposed technical solution. This discussion is omitted in this work for brevity. 

\subsection{Technology Road-map} \label{tech-road-map}

The fit-gap exercise of comparing the existing DCC architecture with our B-SMART reference architecture thus provides us with a number of architectural short-comings (gaps) that can be used to build out a research and development road-map for making the DCC an autonomic smart building.

\begin{figure}
\begin{center}
\includegraphics[width=0.8\textwidth]{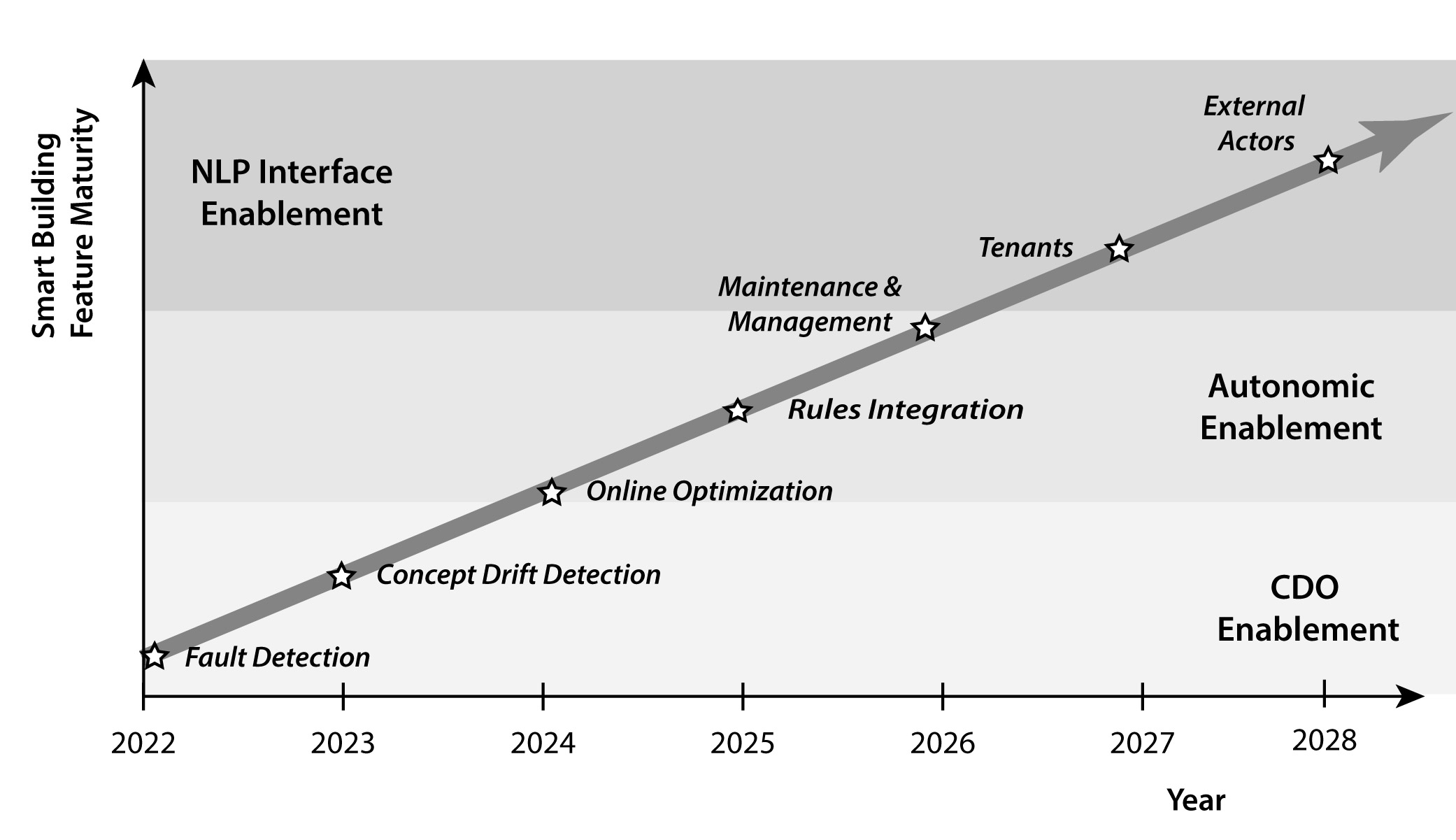}
\end{center}
\caption{\label{road_map} The proposed research and development road map for the DCC.}
\end{figure}

Figure ~\ref{road_map} shows the proposed research and development road map that was developed for the DCC based on the results of the fit-gap analysis against the B-SMART relative architecture presented in the previous sub-section. The horizontal axis in the figure represents the passage of time beginning with the year 2022 and spanning the next six years. The vertical axis represents the technological maturity of smart building features that are proposed to be introduced into the DCC to make it 'smarter' and more autonomic. 

Research and development activities  are grouped into three stages:

\begin{enumerate}
\item \textbf{CDO Enablement}. Change detection and optimization capabilities are required to be in place febore autonomic enablement can begin, because they are used to trigger state transitions (see Figure ~\ref{state-transitions}). Therefore, the initial focus has to be on these activities.
\item \textbf{Autonomic Enablement}. This stage focuses on resolving the gaps and fully automating the control loop shown in Figure ~\ref{ctrl-loop} and state transitions shown in Figure ~\ref{statediag}.
\item \textbf{NLP Enablement}. NLP capabilites are not required to be in place before the previous two stages can proceed, because API-based interfacing capabilities can be used initially. Some of the NLP activities can proceed in parallel with CDO Enablement and Autonomic Enablement activities. The bulk of the NLP enablement activities should proceed, though, once the Autonomic Enablement activities reach sufficient level of technical maturity, because NLP-based interfacing will need to be coordinated by autonomic rules and algorithms.
\end{enumerate}

The first gaps to be addressed would have to be those related to RT Analytics and Batch Analytics components, as part of the CDO Enablement stage. Work focusing on real-time fault detection is currently in progress to help close this gap \citep{mokhtari2022}, and it is anticipated that it will reach technical maturity in 2023. To complement fault detection, concept drift detection needs to be addressed as well. It is proposed to do so in 2024.

Once the RT Analytics and Batch Analytics components have been fully implemented and are capable of supporting automated fault and real-time concept drift detection, work to implement autonomic control loop transition gaps can begin. It is proposed to do so in 2024 and 2025, with projects focusing in on-line energy optimization algorithms and rules engine integration.

Works focusing on NLP enablement of the DCC can begin relatively early with a project to put an NLP interface in front of an existing DCC BAS to complement the existing Web-based interface used by the building maintenance staff. Subsequent projects could extend this approach to interface with the DCC management staff, tenants, vehicles, smart grids, other smart buildings, and other AIs. These projects can be staged over a period of 5 years, with the bulk of the development activities taking place after 2025, to benefit from the rules integration that will be implemented as part of the Autonomic Enablement stage.

The DCC BKR will be enhanced iteratively with features and technologies as the above project proceed. Enhancements may include, for example, introducing a database optimized for data warehousing queries, such as Apache Hive.

\section{Discussion} \label{discussion}

Smart buildings, and by extension smart cities, represent a new and rapidly evolving field. Smart buildings are generally designed and developed by multi-disciplinary teams that include the more traditional engineers and architects, computer scientists, and IT personnel. Although the rapidly advancing field of AI presents great opportunities for enabling smart buildings with intellect, today little exists to help guide these teams in this endeavor.

The B-SMART component model presents a proven, repeatable pattern that demonstrates how to combine AI technologies and big data into a coherent autonomic solution. The layering of the B-SMART architecture encourages different vendors to build interoperable solutions. Systems integrators working on implementing smart buildings should be better able to mix and match products from different vendors to suit their building requirements. This can bring additional benefits when we consider smart cities. For example, basing the Interfacing layer for all smart buildings on NLP will eventually enable smart builds in a smart city to literally talk to each other, as well as to their human occupants and maintenance personnel, in a manner that can be readily understood by humans who don’t have IT training.

The main limitation of this study is related to the general level of maturity of the smart builds field. This field is still very young and is far from maturity. There are very few documented examples of autonomic smart builds. Despite this limitation, B-SMART represents a significant advance in this area and will stimulate both technological advances and architectural maturation of this important field of engineering.

Leveraging B-SMART to speed up the architectural design cycle for a smart building will reduce the costs associated with implementing a smart building, will improve the success rate for these projects, and will reduce resistance to converting traditional buildings to smart buildings. It will produce smart building designs that are well liked by building occupants and are able to evolve and improve by iterative introduction of new technologies over time.

\section{Conclusions} \label{conclusions}

In this paper we presented the Building Systems Management Autonomic Reference Template (B-SMART) - the first reference architecture for autonomic smart buildings. The B-SMART reference architecture is designed to guide the development of new generation of autonomic building automation systems for smart buildings. 

B-SMART can dramatically reduce the duration and cost of the IT and software design cycle for smart buildings by providing IT architects and software developers working on smart builds with a clear pattern to follow.

Our reference architecture explains how to apply a wide range of AI, IoT, and big data technologies to construct a coherent autonomic smart building solution. It encourages decoupling of functionality and independent evolution for key technologies used to construct smart buildings, and encourages standardization and interoperability among smart buildings, smart grids, and smart cities.

B-SMART is designed to support the critically important SOCx processes. This contributes to the ongoing discourse within the AEC community regarding Smart Building performance optimization, providing a framework for the development of future SOCx systems to optimize and maintain building performance.

We also presented an example application of B-SMART by developing a technology road map for the DCC smart building of the Toronto Metropolitan University campus. This road map will serve as a guide for research and development activities focusing on DCC for the next five years.



\bibliographystyle{elsarticle-harv} 
\bibliography{b-smart}





\end{document}